\newif\ifdraft
\drafttrue
\draftfalse

\pdfoutput=1

\documentclass[11pt]{article}

\ifdraft
    \usepackage[review]{ACL2023}

\else
    \usepackage{ACL2023}

\fi

\usepackage{xspace}
\usepackage{amsmath}
\usepackage{amssymb}
\usepackage{amsfonts}
\usepackage{times}
\usepackage{latexsym}
\usepackage[T1]{fontenc}
\usepackage[utf8]{inputenc}
\usepackage{microtype}
\usepackage{inconsolata}
\usepackage{graphicx}
\usepackage{subcaption}
\usepackage{tikz,pgfplots}
\usepackage{booktabs}
\usepackage{cleveref}
\usepackage{multirow}
\usepackage{enumitem}

\graphicspath{{figures/}}
\newcommand{\pretssel}{PRETSSEL\xspace}
\newcommand{\dinopretssel}{DINO-PRETSSEL\xspace}

\title{Textless Acoustic Model with Self-Supervised Distillation for\\Noise-Robust Expressive Speech-to-Speech Translation}

\author{
    Min-Jae Hwang,
    Ilia Kulikov,
    \bf{Benjamin Peloquin},
    \bf{Hongyu Gong},\\
    \bf{Peng-Jen Chen},
    \and
    \bf{Ann Lee}\\
    Meta AI \\
    \texttt{mjhwang@meta.com}
}

\begin{document}
\maketitle
\begin{abstract}
In this paper, we propose a textless acoustic model with a self-supervised distillation strategy for noise-robust expressive speech-to-speech translation (S2ST).
Recently proposed expressive S2ST systems have achieved impressive expressivity preservation performances by cascading unit-to-speech (U2S) generator to the speech-to-unit translation model. 
However, these systems are vulnerable to the presence of noise in input speech, which is an assumption in real-world translation scenarios. 
To address this limitation, we propose a U2S generator that incorporates a distillation with no label (DINO) self-supervised training strategy into it's pretraining process.
Because the proposed method captures noise-agnostic expressivity representation, it can generate qualified speech even in noisy environment.
Objective and subjective evaluation results verified that the proposed method significantly improved the performance of the expressive S2ST system in noisy environments while maintaining competitive performance in clean environments\footnote{
Audio samples are available at\\\url{https://facebookresearch.github.io/seamless_communication/demo/dino_pretssel/index.html}}.
\end{abstract}

\section{Introduction}
Speech-to-speech translation (S2ST), which translates speech in one language into speech in other language is indispensable technique for breaking down language barriers in naturalistic communication among international communities \citep{lavie1997janus, nakamura2006the, wahlster2013verbmobil}. 
Recently, the direct speech-to-unit translation (S2UT) approach that translates source speech into a discretized semantic unit of target speech has been gaining a lot of attention~\citep{lee-etal-2022-direct, lee-etal-2022-textless, chen-etal-2023-speech, inaguma2022unity, SeamlessM4TArXiv}.
Thanks to their ability in modeling semantic units using a single network and the success of large-scale pre-training and data augmentation~\citep{popuri2022enhanced}, their latest models can achieve state-of-the-art translation quality~\citep{inaguma2022unity, SeamlessM4TArXiv}.

On the other hand, it is also important to preserve source speech's expressivity features\footnote{In this work, we define \textit{expressivity} as speech's utterance-level styles such as vocal style, emotion, or tone.} such as vocal style, emotion, or tone during translation process to realize natural conversation with speech translator.
However, it is non-trivial for a direct S2UT system to preserve the expressivity due to its design.
Specifically, since the target discrete units contain linguistic information besides expressivity information~\citep{SeamlessM4TArXiv}, the expressivity of source speech is hard to be captured by S2UT model.
As a results, the translated speech provides monotonic and robotic sound where the source speech's expressivity doesn't exist.

To achieve an expressivity-preserved S2ST system, several studies propose to cascade an additional unit-to-speech (U2S) generator, also known as a \textit{textless acoustic model}, on top of the S2UT model.
This approach proposes the U2S model by replacing the input of the acoustic model used in TTS systems from phonemes to discrete units.
For instance, recently proposed \pretssel-based S2ST system~\citep{barrault2023seamless} generates speech by receiving target language's discrete units and source language speech's expressivity embedding.
The S2ST system with \pretssel can achieve high-quality cross-lingual expressivity transfer and content translation performance because it effectively disentangles the linguistic and paralinguistic information using discrete units and expressivity embedding, respectively.
However, this expressive S2ST framework exhibit issues when applied in the real-world translation scenarios, where the recording environment is noisy.
Note that the expressivity encoder embedding space is trained to represent all information except linguistic one; that means a channel information such as background noise also exists in expressivity embedding along with vocal style, emotion, or tone information. 
Consequently, when the input speech is recorded in the noisy environment, the U2S model tries to transfer background noise to output speech, which critically affects both contents and expressivity preservation performances.

To address aforementioned problem, we propose a textless acoustic model that utilizes the self-distillation with no label (DINO) strategy to its pretraining~\citep{caron2021emerging}.
Following the success of self-supervised speaker representation learning~\citep{chen2023comprehensive}, the proposed method introduces two teacher-student encoders and optimizes these encoders using a self-distillation training strategy. 
Specifically, the student encoder is updated to minimize its output probabilistic distance to the teacher encoder's output. 
Then, the teacher encoder weights are iteratively updated by the exponential moving averaged (EMA) weights of the student encoder. 
In addition, random noise augmentation is applied to the input of both expressivity encoders to learn noise-agnostic expressivity representations.
 
We applied the proposed training strategy to the \pretssel U2S generator, which we refer to as \textbf{\dinopretssel}. 
Experimental results verified that the expressive S2ST system with \dinopretssel outperformed conventional S2ST models in noisy recording environments while still achieving compatitive performance in clean recording environments. 
Specifically, the objective evaluation demonstrated that \dinopretssel achieved more noise-robust content and prosody preservation performance than other systems based on their ASR-BLEU~\citep{jia2019direct} and AutoPCP~\citep{barrault2023seamless} scores. 
Additionally, the subjective evaluation confirmed its superior performance in generating natural speech sound with robust vocal style preservation compared to conventional systems through the mean opinion score (MOS) and speaker-MOS (S-MOS) tests.

\section{Related work}
\subsection{Expressive S2ST}
There have been several studies proposing expressive S2ST model by cascading U2S generator to the S2UT model.
For instance, \pretssel~\citep{barrault2023seamless} and StyleS2ST~\citep{song2023styles2st} adopted FastSpeech (FS)-style non-autoregressive (NAR) U2S generators~\citep{ren2019fastspeech, ren2021fastspeech2} for expressive S2ST.
It also had been proved that the unit-based VoiceBox is also strong NAR U2S generator~\citep{le2023voicebox, barrault2023seamless}.
On the other hand, PolyVoice~\citep{polyvoice} proposed a similar cascaded S2ST by cascading two language models as S2UT and U2S generator components.
Although previous works have shown impressive performance, they didn't consider expressivity preservation in noisy environments. 
To our knowledge, our work is the first work to propose a noise-robust approach to expressive S2ST.


\subsection{Noise-robust expressive TTS}
As expressive TTS systems are becoming more natural and approaching human-level quality, there is a growing interest in incorporating noise-robustness into these systems in order to use it in the real world scenario.
For instance, \citet{hsu2018hierarchical} proposed to disentangle noise information from noisy speech during training process of Gaussian mixture variational autoencoder.
\citet{swiatkowski2023cross} disentangled noise information by using external denoiser~\citep{isik2020poconet}.
During inference, they used only clean speech components for the clean speech generation.

Our work was mostly inspired by \citet{pankov2023dinovits}. 
This system achieved a robust voice cloning system by using VITS-based U2S generator with DINO strategy.
The main difference of our work is that we focus on the cross-lingual S2ST application, whereas earlier work was applied to monolingual TTS application.


\section{Background: Expressive S2ST with \pretssel}
\pretssel is a unit-based textless acoustic model for expressive S2ST system~\citep{barrault2023seamless}.
Specifically, \pretssel is pretrained to reconstruct 80-dimensional Mel-spectrogram with 10-ms interval of input speech from the deduplicated (or \textit{reduced}) XLS-R units~\citep{babu2021xls} with 10K K-means clustering and the same Mel-spectrograms.

\subsection{Architecture of \pretssel}
\label{sec:pretssel}
The \pretssel is composed of the expressivity encoder and the acoustic model.
First, the expressivity encoder extracts a 512-dimensional expressivity embedding vector containing high-level paralinguistic representations from the input Mel-spectrograms.
Specifically, it adopts the variants of ECAPA-TDNN architecture~\citep{desplanques-ecapa-tdnn-2020} that replaces batch normalization~\citep{ioffe2015batch} with layer normalization~\citep{lei2016layer}.

For a pair of expressivity embedding and discrete XLS-R units, the acoustic model generates Mel-spectrograms of output speech.
The acoustic model architecture is based on FS2 architecture~\citep{ren2021fastspeech2} consisting of a series of feed-forward Transformer (FFT) blocks, local prosody predictors, variance adaptors, and decoder FFT blocks.
Major differences are (1) \pretssel uses FiLM conditioning layer \citep{perez2018film, boris2018tadam} to effectively utilize the expressivity embedding, (2) it uses the separately predicted unit duration from external S2UT model, and (3) it individually predicts the binary voiced/unvoiced (VUV) flag and the continuous F0 contour.

\subsection{Pretraining}
During pretraining, expressivity encoder and acoustic model are jointly trained to minimize three loss terms:
\begin{align}
    \mathcal{L}_{pretssel} = \mathcal{L}_{mel} + \lambda_{l} \cdot \mathcal{L}_{local} + \lambda_{f} \cdot \mathcal{L}_{film},\label{eq:pretssel_loss}
\end{align}
where $\mathcal{L}_{mel}$, $\mathcal{L}_{local}$, and $\mathcal{L}_{film}$ denote Mel-spectrogram prediction loss, local prosody prediction loss, and L2 regularization loss at the FiLM layer~\citep{boris2018tadam}, respectively; $\lambda_{v}$ and $\lambda_{v}$ denote weight terms for $\mathcal{L}_{local}$ and $\mathcal{L}_{film}$, respectively.
Specifically, Mel-spectrogram prediction loss is defined by summation of L1 and L2 losses for the predicted Mel-spectrograms before and after PostNet.
In addition, local prosody prediction loss is defined by summation of L2 losses for the continuous F0 and energy contours in logarithm scale, and binary cross entropy (BCE) loss for VUV flag.

\subsection{Application in Expressive S2ST}
\label{sec:pretssel:inference}
For the accurate prediction of translated units and their duration, the work of \pretssel proposes a Prosody UnitY2 S2UT model, which is an expressivity variant of the latest SeamlessM4T V2 model~\citep{barrault2023seamless}.
This S2UT model predicts \textit{original} XLS-R units at 20-ms interval conditioned by \pretssel's expressivity embedding vector.
After predicting original units, the reduced units and their duration are obtained by deduplication process.
Then, \pretssel's acoustic model generates output Mel-spectrograms by taking reduced units, unit durations, expressivity embedding, and target language embedding.
Finally, the HiFi-GAN vocoder~\citep{kong2020hifi} converts Mel-spectrograms into the speech waveform.

Since linguistic information dominates in the discrete units of speech~\citep{SeamlessM4TArXiv}, the expressivity embedding learns paralinguistic information such as prosody, vocal, or channel information as mentioned by~\citet{skerry2018towards}.
This property enables to efficiently transfer the expressivity of source speech into translated speech, especially when the speech is recorded in clean environment.
However, when it comes to the expressive S2ST in real-world situation that assumes noisy condition, the model tries to transfer noise components as well, which critically decrease the quality of translated speech signal.

\section{DINO-PRETSSEL}
\label{sec:dino_pretssel}
\begin{figure}[t]
    \centering
    \includegraphics[width=1.05\linewidth]{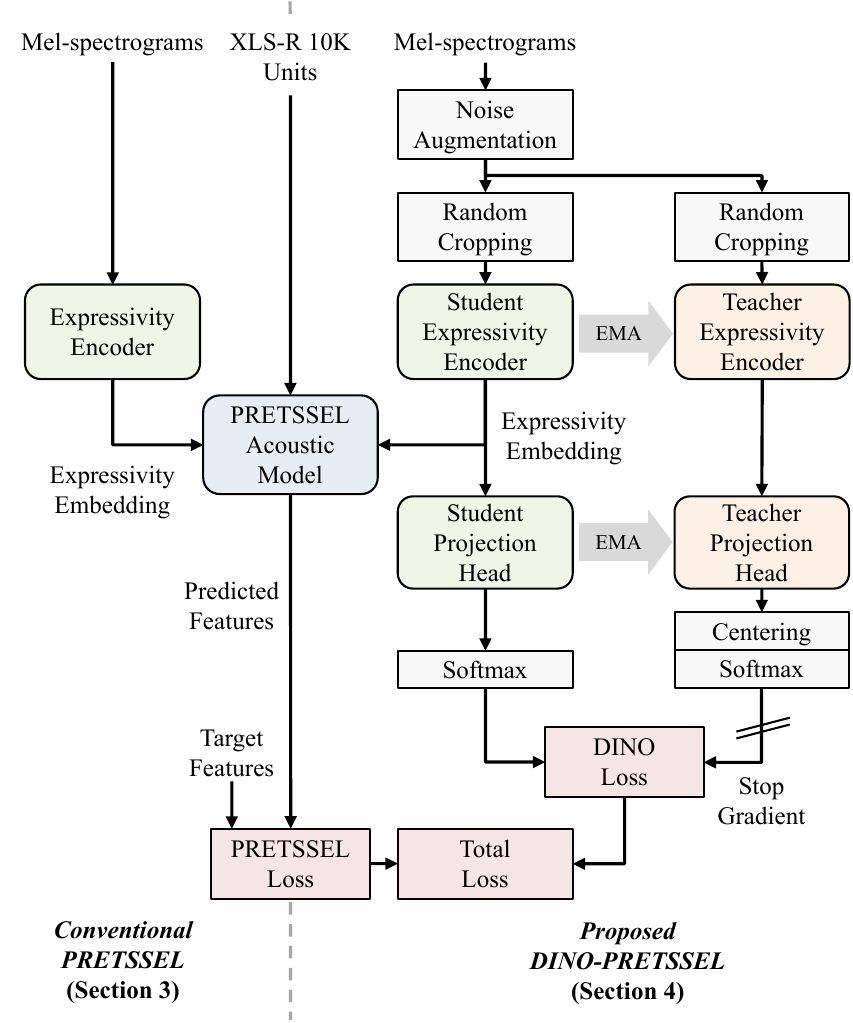}
    \caption{
        Pretraining of conventional PRETSSEL (left) and proposed DINO-PRETSSEL (right).
        Target and predicted features include Mel-spectrograms, pitch, energy, and voicing flag.
    }
    \label{fig:dino_p2v}
\end{figure}
In this section, we propose a \dinopretssel, where the DINO-based self-supervised training strategy is incorporated into \pretssel pretraining as illustrated in \Cref{fig:dino_p2v}.
The details of this process are described in the following sections.

\subsection{Learning robust expressivity embedding with self-distillation}
\dinopretssel utilizes two expressivity encoders termed teacher and student sharing the same ECAPA-TDNN architectures~\citep{desplanques-ecapa-tdnn-2020}.
Unlike the original \pretssel's expressivity encoder, both teacher and student encoder contain additional projection head layers followed by softmax layer to measure probabilistic distance between student and teacher predictions.
The projection heads are multi-layer perceptron (MLP) with linear output layer interleaved by GeLU activation~\citep{hendrycks2016gaussian}.
Then, the student and teacher encoders are iteratively updated following the self-distillation framework.

\paragraph{Student training}
Let $\mathbf{q}_{t}$ and $\mathbf{q}_s$ be the $K$-dimensional outputs obtained by teacher and student encoders, respectively.
Then, we obtain the probability distribution of student output $\mathbf{p}_s$ by applying softmax function as follows:
\begin{equation}
    p_s^{i} = \frac{\text{exp}\left(q_s^i / \tau_s\right)}{\sum_{k=1}^{K}\text{exp}\left(q_s^k / \tau_s\right)},
\end{equation}
where $i$ is the $i^{th}$ dimension of $\mathbf{p}_s$ and $\tau_s$ is the temperature parameter that controls the sharpness of the output distribution.
We apply similar formula to teacher output $\mathbf{p}_t$ by using temperature $\tau_t$.
Then, we train the student encoder to match it's output distribution to the teacher encoder output by minimizing cross-entropy (CE) loss between $\mathbf{p}_s$ and $\mathbf{p}_t$.
To freeze teacher encoder weights, we apply stop gradient operator to $\mathbf{p}_t$.

Following \citet{chen2023comprehensive}, we adopt multi-crop strategy~\citep{caron2020unsupervised} to DINO loss.
Specifically, we randomly sample $L$ long segments and $M$ short segment from single utterance to extract the expressivity embeddings containing long-term and short-term expressivity context. 
All the $L + M$ segments are fed to student encoder, but only $L$ long segments are fed to teacher encoder. 
Then, we compute DINO loss as the combination of CE losses between expressivity embeddings obtained from different segments as follows:
\begin{equation}
    \mathcal{L}_{dino} = \frac{1}{L\cdot(L + M - 1)}\sum_{l=1}^{L}\sum\limits_{\substack{m=1\\m\neq l}}^{L+M} \text{CE}\left(\mathbf{p}_t^l, \mathbf{p}_s^m\right),
\end{equation}
where $l$ and $m$ denote the $l^{th}$ short segment and $m^{th}$ long segment, respectively.

\paragraph{Teacher training}
After updating the student encoder weights $\{\theta_s\}$ by one iteration, teacher network weights $\{\theta_t\}$ are assigned a running average of past student encoder weights by the EMA rule as follows:
\begin{equation}
    \label{eq:teacher_enc}
    \theta_t \leftarrow \lambda_{ema} \cdot \theta_s + (1 - \lambda_{ema}) \cdot \theta_s,
\end{equation}
where $\lambda_{ema}$ controls the extent to which the current student encoder's weights affect the update of the teacher encoder's weights.
Following original DINO study~\citep{caron2021emerging} we gradually increases $\lambda$ from 0.996 to 1.0 until the end of model training using cosine scheduler~\citep{grill2020bootstrap}.


\paragraph{Avoiding model collapse.}
Because the teacher encoder is learned  from the past weights of student encoder, a trivial solution that the encoders can learn is for the teacher to always present uniformly random values or deterministic values by outputting uniformly distributed or single dimension-dominated $\mathbf{p}_t$, respectively.
DINO framework prevents this solution by applying centering and sharpening operations to teacher output distribution.

In detail, the centering operation normalizes logits of softmax distribution by the mean statistic $\mathbf{c}$, which is updated by EMA rule as follows: 
\begin{gather}
    \mathbf{q}_t \leftarrow \mathbf{q}_t - \mathbf{c}, \nonumber \\
    \mathbf{c}   \leftarrow m \cdot \mathbf{c} + (1 - m) \cdot \frac{1}{B}\sum_{i=1}^{B}\mathbf{q_t},
\end{gather}
where $m$ and $B$ denote the momentum factor for EMA update, and batch size, respectively.
The centering operation prevents the situation that one dimension of teacher output dominates other dimensions by normalizing teacher logits to have similar dynamic range.

On the other hand, the sharpening operation makes teacher distribution sharper than the student distribution by setting smaller teacher temperature $\tau_t$ than student temperature $\tau_s$.
Thus, this operation prevents the situation that the teacher distribution to be uniform.

\paragraph{Noise augmentation.}
\label{sec:noiseaug}
To obtain noise-agnostic expressivity representation, we apply random noise augmentation to the input of expressivity encoders.
Then, we train \dinopretssel to predict clean one.
At each iteration, we randomly select a noise signal from the noise database and add it to the input speech using a randomly determined signal-to-noise ratio (SNR).
On the other hand, we always obtain the discrete units for the acoustic model input from the clean speech signal.
By doing so, the model always receives high-quality linguistic input that is not corrupted by noise during training.


\subsection{Pretraining}
We first start from original \pretssel training as described in \Cref{eq:pretssel_loss} to initiate DINO training from the stable states of expressivity encoder.
After obtaining converged expressivity encoder, we apply the DINO framework as detailed in~\Cref{sec:dino_pretssel}.
More specifically, we first obtain teacher and student expressivity encoders from the weights of stabilized expressivity encoder. 
Then, the student encoder is jointly trained along with acoustic model to minimize following loss terms:
\begin{equation}
    \mathcal{L}_{total} = \mathcal{L}_{pretssel} + \lambda_{dino} \cdot \mathcal{L}_{dino},
\end{equation}
where $\lambda_{dino}$ denotes weight term of DINO loss.
Next, the teacher encoder is updated by following~\Cref{eq:teacher_enc}.
This iteration is repeated until DINO loss is converged.

\section{Experimental setting}

\begin{table}
    \small
    \renewcommand{\arraystretch}{0.5}
    \centering
    \begin{tabular}{l|cccc|c}
    \toprule
    \textbf{Language} & \textbf{English}       & \textbf{Spanish} & \textbf{Total} \\ \midrule
    \# utterances ($\times10^3$)  & 10.9            & 4.8           & 24.3    \\
    Duration ($\times10^3$ hrs) & 44.7              & 5.3           & 58.8    \\
    \bottomrule
    \end{tabular}
    \caption{Statistics of \pretssel pretraining datasets per language.
    }
    \label{tables:pretrain_data}
\end{table}

\subsection{Dataset}
We used multi-speaker datasets covering two high-resource languages, i.e., English (En) and Spanish (Es) to train \pretssel models.
We provide a summary of the data statistics, including the number of utterances and duration for each language, in~\Cref{tables:pretrain_data}.
For the evaluation, we used dev and test subsets of mExpresso English to Spanish (En$\rightarrow$Es) and mDRAL Spanish to English (Es$\rightarrow$En) benchmark dataset~\citep{barrault2023seamless}.
More details about evaluation dataset are described in \Cref{sec:appendix:benchmark_data}. 

Note that we could simulate \textit{clean} environment because those mDRAL Es and mExpresso En speeches were recorded in a professional recording studio with minimal background noise.
To simulate \textit{noisy} environment, we obtained random noise signals from DNS-5 dataset~\citep{dubey2023icassp}, and added those signals to source signals by different levels of SNR.

\subsection{Preprocessing}
We extracted XLS-R 10K units, 80-dimensional Mel-spectrograms, continuous F0, VUV flags, and energy.
More details are described in \Cref{sec:appendix:preprocessing}.

\subsection{Architecture}
As described in Section \ref{sec:pretssel}, the architecture of \dinopretssel is similar to the original \pretssel with the exception of the teacher-student expressivity encoders and projection head layers.
The projection head for each expressivity encoders has three fully connected layers with 2,048 hidden dimensions followed by L2 normalization and weight normalization layers~\citep{tim2016weight}.
The output dimension $K$ was set to 65,536.
More details are described in \Cref{sec:appendix:arch}.

\subsection{Training}
We first trained \dinopretssel by $500$k iterations using \pretssel criteria, and fine-tuned another $300$k iterations by using DINO strategy.
For noise augmentation, we randomly selected noise segments in each iteration, and mixed them with the source speech at a 50\% probability with a random SNR ranging from 6dB to 40dB.
We set the length of short and long segments to 4- and 6-seconds, respectively.
We set the number of short and long segments to 4 and 2, respectively.
More details are described in \Cref{sec:appendix:training}.

\subsection{Expressive S2ST inference}
\label{sec:exp:inference}
Firstly, we used the many-to-many version of Prosody UnitY2~\citep{barrault2023seamless} to translate input speech into target language's XLS-R 10K units as described in \Cref{sec:pretssel:inference}.
Then, the \dinopretssel took XLS-R 10K and last 4 seconds of Mel-spectrograms to synthesize the Mel-spectrograms at target language.
Finally, the HiFi-GAN  V1 vocoder~\citep{kong2020hifi} converted Mel-spectrograms into the speech waveform at 24-kHz sampling rate.
Unlike original \pretssel work, we didn't include audio watermarking technique to prevent possible distortion from watermarking.

\subsection{Benchmarking systems}
We included four expressive S2ST systems in the experiments.
Note that in case of \pretssel-based systems, we used the same Prosody UnitY2 and HiFi-GAN models used at original \pretssel work~\citep{barrault2023seamless} for a fair comparison.

\paragraph{S2TT + TTS.}
We combined a SeamlessM4T V2's S2TT module~\citep{barrault2023seamless} and Coqui-XTTS V2 model\footnote{\url{https://huggingface.co/coqui/XTTS-v2}}.
For the expressivity transfer, the source speech of SeamlessM4T V2 was conditioned to Coqui-XTTS V2.

\paragraph{\pretssel.}
We combined a Prosody UnitY2 and \pretssel model as described in~\Cref{sec:pretssel:inference}.

\paragraph{Denoiser + \pretssel.}
We combined a Prosody UnitY2 and \pretssel with high-quality speech enhancement model.
Specifically, we applied MetricGAN+ denoiser\footnote{\url{https://huggingface.co/speechbrain/metricgan-plus-voicebank}}~\citep{fu2021metricgan+} to the input of \pretssel for removing noise components.

\paragraph{Proposed \dinopretssel}
We combined a Prosody UnitY2 and \dinopretssel as described in~\Cref{sec:exp:inference}.

\begin{figure}[t]
	\centering
        \hspace*{-0.5cm}
        \includegraphics[width=1.1\linewidth]{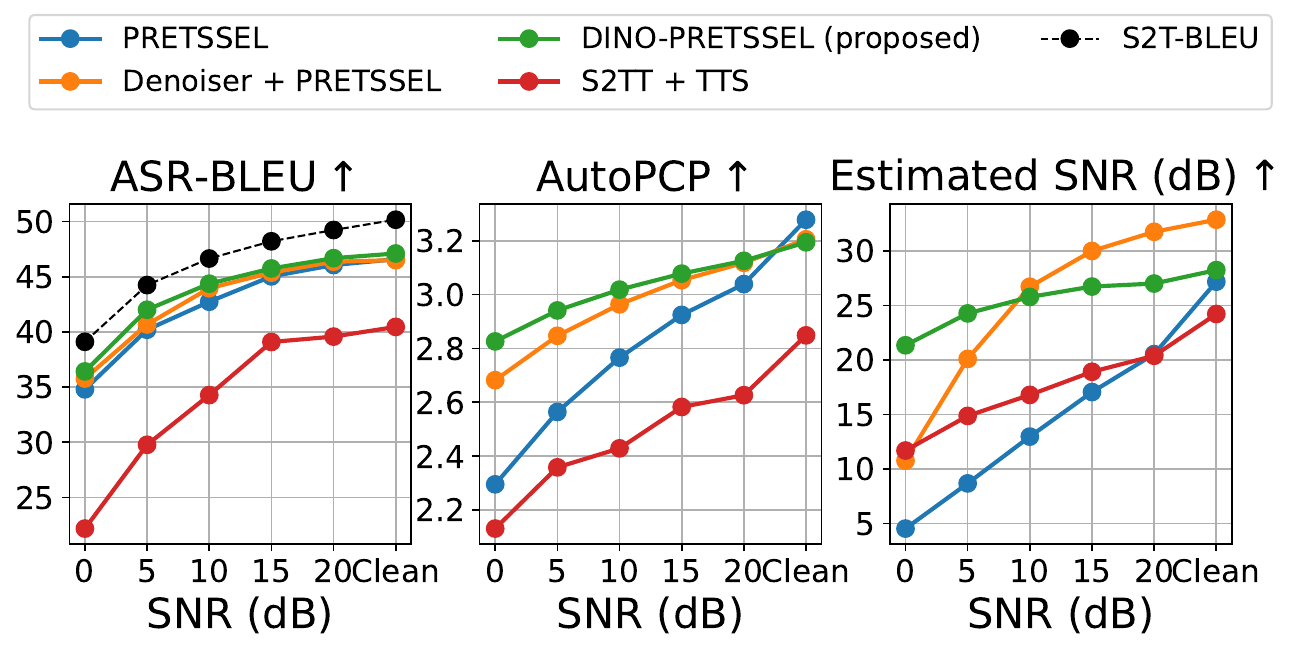}	
	\centering
	\caption{
            Objective evaluation results of various expressive S2ST systems under different SNR conditions.
        }
	\vspace{-0.3cm}
	\label{fig:obj_results-pred_unit}
\end{figure}
\begin{table*}[t]
\small
\setlength{\tabcolsep}{2.5pt}

\centering
\hspace*{-6mm}
\begin{tabular}{cl|cc|cc|cc|cc}
\toprule
\multicolumn{1}{c}{\multirow{3}{*}{Label}} & \multicolumn{1}{c|}{\multirow{3}{*}{System}} & \multicolumn{4}{c|}{mExpresso (En$\rightarrow$Es)} & \multicolumn{4}{c}{mDRAL (Es$\rightarrow$En)} \\ \cmidrule{3-10} 
 & \multicolumn{1}{c|}{} & \multicolumn{2}{c|}{Naturalness MOS$\uparrow$} & \multicolumn{2}{c|}{S-MOS$\uparrow$} & \multicolumn{2}{c|}{Naturalness MOS$\uparrow$} & \multicolumn{2}{c}{S-MOS$\uparrow$} \\\cmidrule{3-10} 
 & \multicolumn{1}{c|}{} & Clean & Noisy & Clean & Noisy & Clean & Noisy & Clean & Noisy \\ \midrule
S1 & S2TT + TTS & \textbf{3.28$\pm$0.14} & 2.96$\pm$0.14 & 3.11$\pm$0.16 & 2.40$\pm$0.16 & 3.11$\pm$0.15 & 2.74$\pm$0.15 & 3.05$\pm$0.17 & 2.49$\pm$0.16 \\
S2 & PRETSSEL & 3.24$\pm$0.14 & 3.02$\pm$0.14 & \textbf{3.58$\pm$0.15} & 2.99$\pm$0.16 & \textbf{3.69$\pm$0.13} & 2.89$\pm$0.16 & \textbf{3.99$\pm$0.14} & 2.88$\pm$0.17 \\
S3 & Denoiser + PRETSSEL & 3.13$\pm$0.14 & 3.49$\pm$0.13 & 3.39$\pm$0.15 & 2.88$\pm$0.15 & 3.54$\pm$0.14 & 3.61$\pm$0.14 & 3.68$\pm$0.15 & 2.88$\pm$0.17 \\
S4 & DINO-PRETSSEL (proposed) & 3.20$\pm$0.14 & \textbf{3.54$\pm$0.13} & 3.26$\pm$0.16 & \textbf{3.02$\pm$0.15} & 3.61$\pm$0.14 & \textbf{3.64$\pm$0.13} & 3.63$\pm$0.16 & \textbf{3.15$\pm$0.17}
\\ \bottomrule
\end{tabular}
\caption{Subjective evaluation results for various expressive S2ST systems with a 95\% confidence interval. The highest scores are in bold typeface. "Clean" and "Noisy" denote that the source speech of S2ST system was clean and noisy, respectively.}
\label{table:subj_eval_results}
\end{table*}

\section{Objective evaluation}
\label{sec:obj_eval}

\subsection{Metrics}
\label{sec:obj_eval:metrics}
In the objective evaluation, we measured ASR-BLEU, AutoPCP, and SNR to measure contents preservation, prosody preservation, and noise suppression performances, respectively.
We also included S2T-BLEU scores to specify the upper-bound of ASR-BLEU score.
We averaged the scores of individual utterances to obtain single representative score.
We detailed the evaluation metrics in \Cref{sec:appendix:obj_eval_metrics}.

\subsection{Results}
\label{sec:obj_eval:results}
We present the objective evaluation results at Figure~\ref{fig:obj_results-pred_unit}. 
In all cases, we confirmed the superior performance of \pretssel-based models compared to the S2TT + TTS baseline. 
We analyze the trends among the \pretssel models as follows.

When the source speech was clean, the noise-robust PRETSSELs (i.e., \dinopretssel and denoiser-based \pretssel) showed a slight drop in prosody preservation performance compared to the original \pretssel (AutoPCP), while they demonstrated a slight improvement in content preservation performance (ASR-BLEU).
As for the noise suppression performance, we found that \dinopretssel showed the closest SNR values to the original \pretssel.
On the other hand, the denoiser-based \pretssel provided the highest SNR value, indicating the lowest noise level. 
We hypothesize that its significantly higher SNR value might indicate that the denoiser removed too much noise information that exists in the source speech, which could potentially degrade the naturalness of the translated speech.
We will analyze this at \Cref{sec:subj_eval}.

When the source speech was corrupted by noise, the proposed \dinopretssel demonstrated the best contents preservation (ASR-BLEU), prosody preservation (AutoPCP), and noise suppression (SNR) performance regardless of input noise level.
One interesting observation was that the proposed \dinopretssel model remained robustness to the noise, even when it was much stronger than the noise used during training, i.e., less than 6dB SNR.
We presented full results in \Cref{sec:appendix:detailed_obj_eval_scores}.

\section{Subjective evaluation}
\label{sec:subj_eval}

\subsection{Metrics}
For the subjective evaluation, we conducted MOS test to measure the naturalness of translated speeches.
We also conducted S-MOS test that measures a vocal style similarity between source and translated speeches.
We detailed the evaluation metrics in \Cref{sec:appendix:subj_eval_metrics}.

\subsection{Results}
\label{sec:subj_eval:results}

The subjective evaluation results are shown in \Cref{table:subj_eval_results}. 
We analyze the trends as follows.

\paragraph{Naturalness.}
When the source speech was clean, \dinopretssel provided slightly lower naturalness than the original \pretssel (clean source; S4 vs. S2). 
This is mainly because DINO-PRETSSEL’s training objective is broader than PRETSSEL’s objective. 
More specifically, DINO-PRETSSEL needs to learn both denoising and expressivity preservation, whereas PRETSSEL only learns expressivity preservation. 
Because the clean speech translation case doesn’t consider denoising ability, DINO-PRETSSEL’s performance can be worse than PRETSSEL. 
This could be also interpreted as a trade-off for gaining noise-robustness at the expense of performance in clean environments. 
However, \dinopretssel still demonstrated higher naturalness than denoiser-based \pretssel (clean source; S4 vs. S3). 
This supported our hypothesis in \Cref{sec:obj_eval:results} that removing too much noise with the denoiser could have a negative impact on the naturalness of the translated speech.

When the source speech became noisy, in contrast to other models, the naturalness performance of noise-robust \pretssel models even increased (noisy sources; S3 and S4). 
Specifically, \dinopretssel proved to be able to generate the most natural speech compared to other systems in noisy environments (noisy source; S4 vs. others).
More specifically, in noisy environments, \dinopretssel achieved 3.54 and 3.64 MOS results, respectively 0.52 and 0.75 scores higher than the original \pretssel in the mExpresso (En$\rightarrow$Es) and mDRAL (Es$\rightarrow$Es) subsets.

\paragraph{Vocal style preservation.}
When the source speech was clean, the original \pretssel showed the highest performance (clean source; S2 vs. others).
In contrast, \dinopretssel showed the lowest vocal style preservation performance among other \pretssel-based models (clean source; S4 vs. S2 and S3).
This trend was similar to the results observed at MOS test, that means there was a trade-off for gaining noise-robustness.

However, when the source speech became noisy, \dinopretssel demonstrated outperforming robustness in vocal style preservation by achieving higher S-MOS results compared to other systems (noisy source; S4 vs. others).
More specifically, \dinopretssel achieved 3.02 and 3.15 S-MOS results in noisy environments, respectively 0.03 and 0.27 scores higher than the original \pretssel in the mExpresso (En$\rightarrow$Es) and mDRAL (Es$\rightarrow$Es) subsets.

\section{Expressive S2ST in-the-wild}
\begin{table}[t]
    \small
    \setlength\tabcolsep{3.5pt}
    \centering
    \hspace*{-3mm}
    \begin{tabular}{cl|c|c} \toprule
    Label & \multicolumn{1}{c|}{System} & MOS$\uparrow$ & S-MOS$\uparrow$ \\ \midrule
    S2 & PRETSSEL & 3.01$\pm$0.16 & 2.63$\pm$0.17 \\
    S3 & Denoiser + PRETSSEL & 3.43$\pm$0.13 & 2.51$\pm$0.16 \\
    S4 & DINO-PRETSSEL (proposed) & \textbf{3.59$\pm$0.13} & \textbf{3.11$\pm$0.16}
    \\ \bottomrule
    \end{tabular}
    \caption{
        Subjective evaluation results for the En$\rightarrow$Es translation of real-world noisy data with 95\% confident interval.
        The highest scores are in bold typeface.
    }
    \label{table:human_eval_realworld}
\end{table}

To test the robustness of \dinopretssel in real-world noisy recording environments, we conducted a subjective evaluation using noisy samples of the VoxLingua107 dataset~\citep{valk2021slt}, which were collected from En videos on YouTube.
When choosing source speech, we first measured SNR, and randomly selected 100 noisy speeches having SNR values from 5dB to 15dB.
We then applied Silero voice activity detection~\citep{SileroVAD} to remove leading and trailing silence.
Then, we conducted MOS and S-MOS tests for the En$\rightarrow$Es translation as shown the results in~\Cref{table:human_eval_realworld}.
The evaluation results verified the proposed \dinopretssel presented significantly higher performance to the noisy samples, especially compared to both \pretssel and denoiser-based \pretssel.
Specifically, it achieved 3.59 MOS and 3.11 S-MOS results, which were 0.58 and 0.48 scores higher than those of original \pretssel, respectively.


\section{Ablation study}
\subsection{Visualizing expressivity embeddings}


The expressivity encoder of proposed \dinopretssel can effectively extract consistent vocal style and prosody information from source speech even in noisy environment.
To show this, we extracted expressivity embeddings from clean and noisy speeches, and drew the t-distributed stochastic neighbor embedding (t-SNE) plots~\cite{van2008visualizing}.
In particular, we chose confused, happy and sad speech samples from Expresso En benchmarking data~\citep{barrault2023seamless}, which showed clear differences in their speaking style.
To simulate noisy environment, we randomly obtained noise from DNS-5 dataset~\citep{dubey2023icassp}, and mixed those with the speeches by 10dB SNR.
When drawing the t-SNE plot, we marked clusters by following \{speaker ID, style ID\} labels. 

\begin{figure}[t]
	\centering
	\begin{subfigure}[b]{0.48\columnwidth}
		\centering
	    \includegraphics[width=0.99\columnwidth]{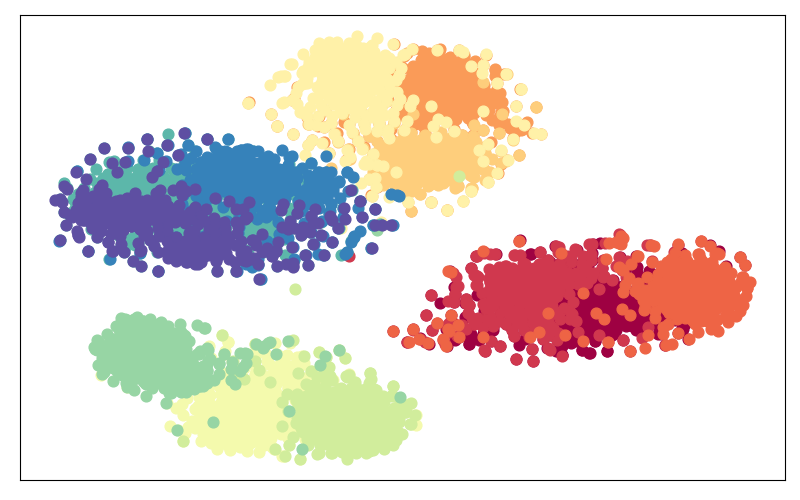}
		\caption{\pretssel.}
   		\label{fig:tsne_ori_clean}
   	\end{subfigure}%
        \begin{subfigure}[b]{0.48\columnwidth}
		\centering
	    \includegraphics[width=0.99\columnwidth]{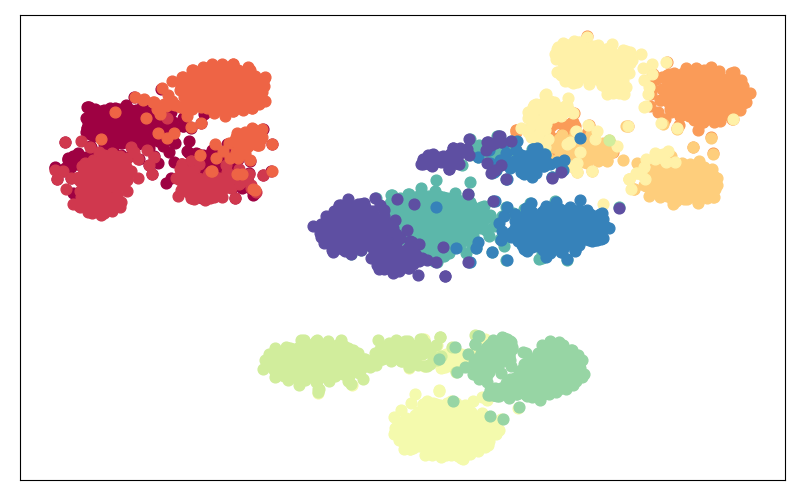}     
		\caption{\dinopretssel.}
   		\label{fig:tsne_dino_clean}
   	\end{subfigure}%
	\centering
        \caption{t-SNE plot of expressivity embeddings obtained from clean speeches.}
	\label{fig:tsne_plot_ori}
\end{figure}

\begin{figure}[t]
	\centering
	\begin{subfigure}[b]{0.48\columnwidth}
		\centering
	    \includegraphics[width=0.99\columnwidth]{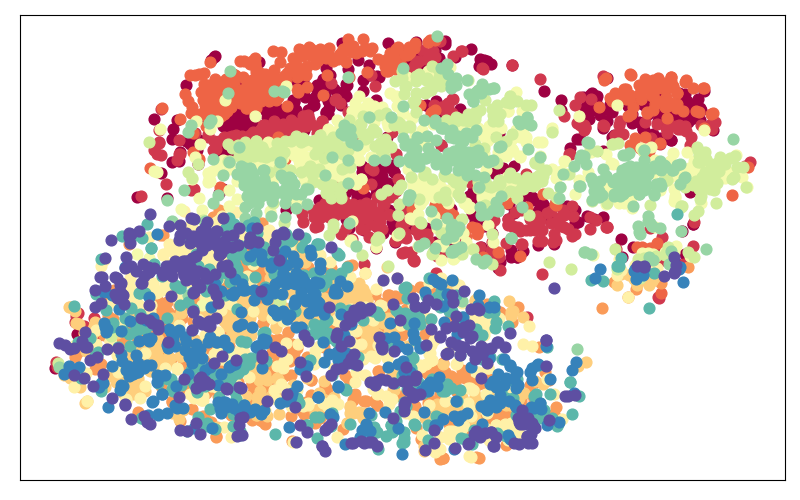}
		\caption{\pretssel.}
   		\label{fig:tsne_ori_noisy}
   	\end{subfigure}
	\begin{subfigure}[b]{0.48\columnwidth}
		\centering
	    \includegraphics[width=0.99\columnwidth]{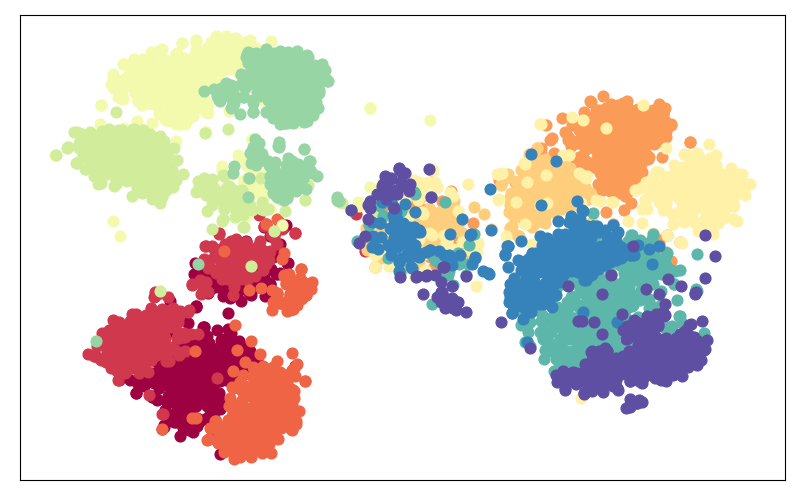}
		\caption{\dinopretssel.}
   		\label{fig:tsne_dino_noisy}
   	\end{subfigure}
	\centering
        \caption{t-SNE plot of expressivity embeddings obtained from noisy speeches.}
	\label{fig:tsne_plot_dino}
\end{figure}

As illustrated in \Cref{fig:tsne_plot_ori}, both expressivity embeddings from \pretssel and \dinopretssel could be distinguishable by following the speaker and style labels when the source speech was clean.
However, as illustrated in \Cref{fig:tsne_plot_dino}, the \pretssel lost its ability to distinguish speaker and style, whereas the \dinopretssel still preserved the clusters when the input speech was corrupted by noise.
This verified the robustness of proposed \dinopretssel's expressivity encoder in the noisy environment.

\subsection{Objective evaluation using ground-truth units}
\begin{figure}[t]
	\centering
        \hspace*{-0.3cm}
        \includegraphics[width=1.1\linewidth]{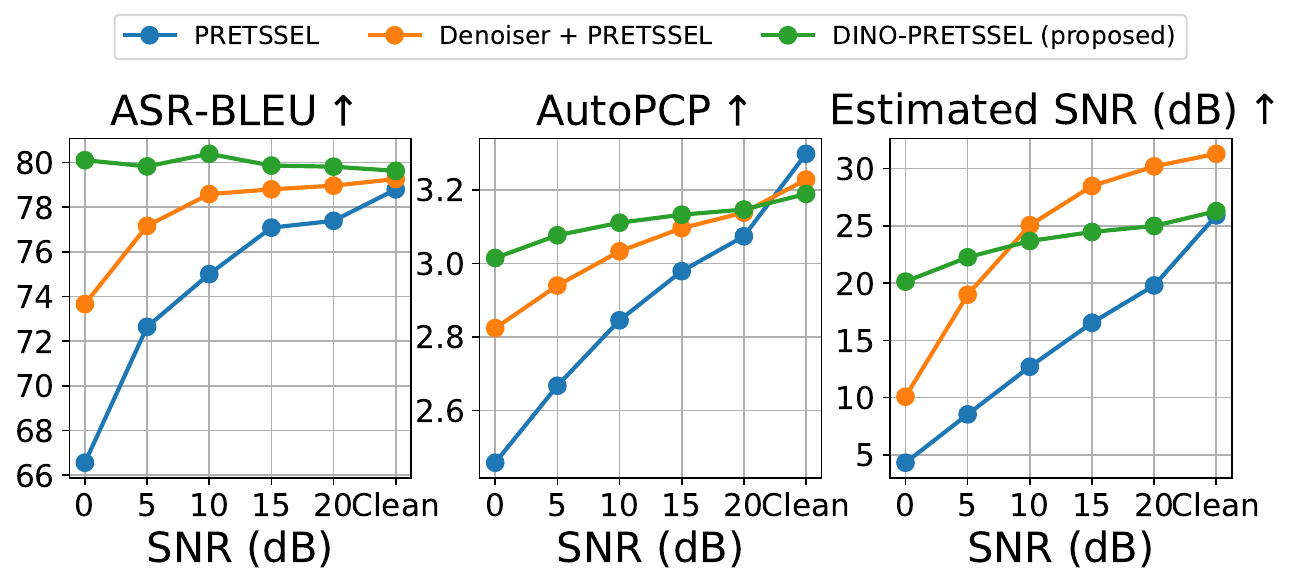}	
	\caption{
            Objective evaluation results of various \pretssel models with ground-truth units. 
        }
	\label{fig:obj_results-gold_unit}
\end{figure}

By using a ground-truth unit as input, we can simulate the best performance of the U2S models when there is no translation error from the preceding S2UT model.
To evaluate the system in this situation, we generated speeches using ground-truth XLS-R 10K units at target language, and measured ASR-BLEU, AutoPCP and SNR metrics.

The evaluation results are presented in \Cref{fig:obj_results-gold_unit}.
In overall, we could observe that the trends were similar to those reported in Section \ref{sec:obj_eval:results}. 
However, there was a noticeable difference in the ASR-BLEU results that \dinopretssel were not affected by the presence of noise in the source speech.
This means that the drop in ASR-BLEU scores for \dinopretssel in the S2ST pipeline was primarily due to translation errors from the S2UT model.
That means, it is possible further improve translation quality of S2ST system by enhancing the robustness of the S2UT model as well, and this could be a potential direction for future work.
Full results are detailed in \Cref{sec:appendix:detailed_obj_eval_scores}.

\section{Conclusion}
In this paper, we proposed \dinopretssel, which incorporated DINO training strategy into the \pretssel-based U2S generator for the noise-robust S2ST system.
As our method employed strong self-distillation method in learning expressivity representation, the U2S generator could robustly transfer source speech's expressivity at S2ST in noisy recording environment.
The objective and subjective evaluations conducted on noisy source speech consistently verified that the proposed \dinopretssel outperformed other systems through its high ASR-BLEU, AutoPCP, MOS, and S-MOS results.
We also verified that \dinopretssel has robustness in real-world noisy speech. 
Future works include improving the robustness of Prosody UnitY2 by utilzing this robust expressivity embedding. 
Additionally, we plan to explore additional data augmentation methods such as reverberation for the training of \dinopretssel.

\section{Limitations}
The DINO strategy significantly reduces the pretraining speed because it requires extracting expressivity embeddings by multiple times, and the head layers also increases computational complexity.
In our experiments, \dinopretssel took 12.7 days for pretraining, which is 5 days longer than \pretssel's 7.9 days.

In addition, because the proposed system transfers source speech's expressivity, it has potential risk of abusing biometric data of source speech.
However, this risk can be mitigated by adopting audio watermarking technique to the translated speech waveforms~\citep{roman2024proactive}.




\bibliography{anthology,custom}

\begin{thebibliography}{49}
\expandafter\ifx\csname natexlab\endcsname\relax\def\natexlab#1{#1}\fi

\bibitem[{Ba et~al.(2016)Ba, Kiros, and Hinton}]{lei2016layer}
Lei~Jimmy Ba, Jamie~Ryan Kiros, and Geoffrey~E. Hinton. 2016.
\newblock Layer normalization.
\newblock \emph{CoRR}, abs/1607.06450.

\bibitem[{Babu et~al.(2022)Babu, Wang, Tjandra, Lakhotia, Xu, Goyal, Singh, {von Platen}, Saraf, Pino, Baevski, Conneau, and Auli}]{babu2021xls}
Arun Babu, Changhan Wang, Andros Tjandra, Kushal Lakhotia, Qiantong Xu, Naman Goyal, Kritika Singh, Patrick {von Platen}, Yatharth Saraf, Juan Pino, Alexei Baevski, Alexis Conneau, and Michael Auli. 2022.
\newblock \href {https://doi.org/10.21437/Interspeech.2022-143} {{XLS-R: Self-supervised Cross-lingual Speech Representation Learning at Scale}}.
\newblock In \emph{Proc. Interspeech 2022}, pages 2278--2282.

\bibitem[{Caron et~al.(2020)Caron, Misra, Mairal, Goyal, Bojanowski, and Joulin}]{caron2020unsupervised}
Mathilde Caron, Ishan Misra, Julien Mairal, Priya Goyal, Piotr Bojanowski, and Armand Joulin. 2020.
\newblock Unsupervised learning of visual features by contrasting cluster assignments.
\newblock In \emph{Proceedings of the 34th International Conference on Neural Information Processing Systems}.

\bibitem[{Caron et~al.(2021)Caron, Touvron, Misra, J\'egou, Mairal, Bojanowski, and Joulin}]{caron2021emerging}
Mathilde Caron, Hugo Touvron, Ishan Misra, Herv\'e J\'egou, Julien Mairal, Piotr Bojanowski, and Armand Joulin. 2021.
\newblock Emerging properties in self-supervised vision transformers.
\newblock In \emph{Proceedings of the International Conference on Computer Vision (ICCV)}.

\bibitem[{Chen et~al.(2023{\natexlab{a}})Chen, Tran, Yang, Du, Kao, Chung, Tomasello, Duquenne, Schwenk, Gong, Inaguma, Popuri, Wang, Pino, Hsu, and Lee}]{chen-etal-2023-speech}
Peng-Jen Chen, Kevin Tran, Yilin Yang, Jingfei Du, Justine Kao, Yu-An Chung, Paden Tomasello, Paul-Ambroise Duquenne, Holger Schwenk, Hongyu Gong, Hirofumi Inaguma, Sravya Popuri, Changhan Wang, Juan Pino, Wei-Ning Hsu, and Ann Lee. 2023{\natexlab{a}}.
\newblock \href {https://aclanthology.org/2023.findings-acl.307} {Speech-to-speech translation for a real-world unwritten language}.
\newblock In \emph{Findings of the Association for Computational Linguistics: ACL 2023}, pages 4969--4983, Toronto, Canada. Association for Computational Linguistics.

\bibitem[{Chen et~al.(2023{\natexlab{b}})Chen, Qian, Han, Qian, and Zeng}]{chen2023comprehensive}
Zhengyang Chen, Yao Qian, Bing Han, Yanmin Qian, and Michael Zeng. 2023{\natexlab{b}}.
\newblock A comprehensive study on self-supervised distillation for speaker representation learning.
\newblock In \emph{2022 IEEE Spoken Language Technology Workshop (SLT)}, pages 599--604. IEEE.

\bibitem[{Defossez et~al.(2020)Defossez, Synnaeve, and Adi}]{defossez2020real}
Alexandre Defossez, Gabriel Synnaeve, and Yossi Adi. 2020.
\newblock Real time speech enhancement in the waveform domain.
\newblock In \emph{Interspeech}.

\bibitem[{Desplanques et~al.(2020)Desplanques, Thienpondt, and Demuynck}]{desplanques-ecapa-tdnn-2020}
Brecht Desplanques, Jenthe Thienpondt, and Kris Demuynck. 2020.
\newblock \href {https://doi.org/10.21437/Interspeech.2020-2650} {{ECAPA-TDNN:} emphasized channel attention, propagation and aggregation in {TDNN} based speaker verification}.
\newblock In \emph{Interspeech 2020, 21st Annual Conference of the International Speech Communication Association, Virtual Event, Shanghai, China, 25-29 October 2020}, pages 3830--3834. {ISCA}.

\bibitem[{Dong et~al.(2023)Dong, Huang, Tian, Xu, Ko, Zhao, Feng, Li, Wang, Cheng, Yue, Bai, Chen, Lu, Ma, Wang, Wang, and Wang}]{polyvoice}
Qianqian Dong, Zhiying Huang, Qiao Tian, Chen Xu, Tom Ko, Yunlong Zhao, Siyuan Feng, Tang Li, Kexin Wang, Xuxin Cheng, Fengpeng Yue, Ye~Bai, Xi~Chen, Lu~Lu, Zejun Ma, Yuping Wang, Mingxuan Wang, and Yuxuan Wang. 2023.
\newblock \href {https://doi.org/10.48550/arXiv.2306.02982} {Polyvoice: Language models for speech to speech translation}.
\newblock \emph{CoRR}, abs/2306.02982.

\bibitem[{Dubey et~al.(2023)Dubey, Aazami, Gopal, Naderi, Braun, Cutler, Gamper, Golestaneh, and Aichner}]{dubey2023icassp}
Harishchandra Dubey, Ashkan Aazami, Vishak Gopal, Babak Naderi, Sebastian Braun, Ross Cutler, Hannes Gamper, Mehrsa Golestaneh, and Robert Aichner. 2023.
\newblock Icassp 2023 deep noise suppression challenge.
\newblock In \emph{ICASSP}.

\bibitem[{Fu et~al.(2021)Fu, Yu, Hsieh, Plantinga, Ravanelli, Lu, and Tsao}]{fu2021metricgan+}
Szu-Wei Fu, Cheng Yu, Tsun-An Hsieh, Peter Plantinga, Mirco Ravanelli, Xugang Lu, and Yu~Tsao. 2021.
\newblock Metricgan+: An improved version of metricgan for speech enhancement.
\newblock \emph{arXiv preprint arXiv:2104.03538}.

\bibitem[{Grill et~al.(2020)Grill, Strub, Altch{\'e}, Tallec, Richemond, Buchatskaya, Doersch, Avila~Pires, Guo, Gheshlaghi~Azar et~al.}]{grill2020bootstrap}
Jean-Bastien Grill, Florian Strub, Florent Altch{\'e}, Corentin Tallec, Pierre Richemond, Elena Buchatskaya, Carl Doersch, Bernardo Avila~Pires, Zhaohan Guo, Mohammad Gheshlaghi~Azar, et~al. 2020.
\newblock Bootstrap your own latent-a new approach to self-supervised learning.
\newblock \emph{Advances in neural information processing systems}, 33:21271--21284.

\bibitem[{Hendrycks and Gimpel(2016)}]{hendrycks2016gaussian}
Dan Hendrycks and Kevin Gimpel. 2016.
\newblock Gaussian error linear units (gelus).
\newblock \emph{arXiv preprint arXiv:1606.08415}.

\bibitem[{Hermans et~al.(2017)Hermans, Spanakis, and M{\"o}ckel}]{hermans2017accumulated}
Joeri~R Hermans, Gerasimos Spanakis, and Rico M{\"o}ckel. 2017.
\newblock Accumulated gradient normalization.
\newblock In \emph{Asian Conference on Machine Learning}, pages 439--454. PMLR.

\bibitem[{Holtzman et~al.(2019)Holtzman, Buys, Forbes, and Choi}]{ari2019thecurious}
Ari Holtzman, Jan Buys, Maxwell Forbes, and Yejin Choi. 2019.
\newblock The curious case of neural text degeneration.
\newblock \emph{CoRR}, abs/1904.09751.

\bibitem[{Hsu et~al.(2018)Hsu, Zhang, Weiss, Zen, Wu, Wang, Cao, Jia, Chen, Shen et~al.}]{hsu2018hierarchical}
Wei-Ning Hsu, Yu~Zhang, Ron~J Weiss, Heiga Zen, Yonghui Wu, Yuxuan Wang, Yuan Cao, Ye~Jia, Zhifeng Chen, Jonathan Shen, et~al. 2018.
\newblock Hierarchical generative modeling for controllable speech synthesis.
\newblock In \emph{International Conference on Learning Representations}.

\bibitem[{Inaguma et~al.(2022)Inaguma, Popuri, Kulikov, Chen, Wang, Chung, Tang, Lee, Watanabe, and Pino}]{inaguma2022unity}
Hirofumi Inaguma, Sravya Popuri, Ilia Kulikov, Peng-Jen Chen, Changhan Wang, Yu-An Chung, Yun Tang, Ann Lee, Shinji Watanabe, and Juan Pino. 2022.
\newblock Unity: Two-pass direct speech-to-speech translation with discrete units.
\newblock \emph{arXiv preprint arXiv:2212.08055}.

\bibitem[{Ioffe and Szegedy(2015)}]{ioffe2015batch}
Sergey Ioffe and Christian Szegedy. 2015.
\newblock Batch normalization: Accelerating deep network training by reducing internal covariate shift.
\newblock In \emph{Proceedings of the 32nd International Conference on Machine Learning}, volume~37, pages 448--456. PMLR.

\bibitem[{Isik et~al.(2020)Isik, Giri, Phansalkar, Valin, Helwani, and Krishnaswamy}]{isik2020poconet}
Umut Isik, Ritwik Giri, Neerad Phansalkar, Jean-Marc Valin, Karim Helwani, and Arvindh Krishnaswamy. 2020.
\newblock Poconet: Better speech enhancement with frequency-positional embeddings, semi-supervised conversational data, and biased loss.

\bibitem[{{ITU-T Recommendation P.808}(2018)}]{p808}
{ITU-T Recommendation P.808}. 2018.
\newblock {Subjective evaluation of speech quality with a crowdsourcing approach}.

\bibitem[{Jia et~al.(2019)Jia, Weiss, Biadsy, Macherey, Johnson, Chen, and Wu}]{jia2019direct}
Ye~Jia, Ron~J Weiss, Fadi Biadsy, Wolfgang Macherey, Melvin Johnson, Zhifeng Chen, and Yonghui Wu. 2019.
\newblock Direct speech-to-speech translation with a sequence-to-sequence model.
\newblock \emph{Interspeech 2019}.

\bibitem[{Kong et~al.(2020)Kong, Kim, and Bae}]{kong2020hifi}
Jungil Kong, Jaehyeon Kim, and Jaekyoung Bae. 2020.
\newblock Hifi-gan: Generative adversarial networks for efficient and high fidelity speech synthesis.
\newblock \emph{Advances in Neural Information Processing Systems}, 33:17022--17033.

\bibitem[{Lavie et~al.(1997)Lavie, Waibel, Levin, Finke, Gates, Gavalda, Zeppenfeld, and Zhan}]{lavie1997janus}
A.~Lavie, A.~Waibel, L.~Levin, M.~Finke, D.~Gates, M.~Gavalda, T.~Zeppenfeld, and Puming Zhan. 1997.
\newblock Janus-iii: speech-to-speech translation in multiple languages.
\newblock In \emph{1997 IEEE International Conference on Acoustics, Speech, and Signal Processing}, volume~1, pages 99--102 vol.1.

\bibitem[{Le et~al.(2023)Le, Vyas, Shi, Karrer, Sari, Moritz, Williamson, Manohar, Adi, Mahadeokar et~al.}]{le2023voicebox}
Matthew Le, Apoorv Vyas, Bowen Shi, Brian Karrer, Leda Sari, Rashel Moritz, Mary Williamson, Vimal Manohar, Yossi Adi, Jay Mahadeokar, et~al. 2023.
\newblock Voicebox: Text-guided multilingual universal speech generation at scale.
\newblock \emph{arXiv preprint arXiv:2306.15687}.

\bibitem[{Lee et~al.(2022{\natexlab{a}})Lee, Chen, Wang, Gu, Popuri, Ma, Polyak, Adi, He, Tang, Pino, and Hsu}]{lee-etal-2022-direct}
Ann Lee, Peng-Jen Chen, Changhan Wang, Jiatao Gu, Sravya Popuri, Xutai Ma, Adam Polyak, Yossi Adi, Qing He, Yun Tang, Juan Pino, and Wei-Ning Hsu. 2022{\natexlab{a}}.
\newblock \href {https://doi.org/10.18653/v1/2022.acl-long.235} {Direct speech-to-speech translation with discrete units}.
\newblock In \emph{Proceedings of the 60th Annual Meeting of the Association for Computational Linguistics (Volume 1: Long Papers)}, pages 3327--3339, Dublin, Ireland. Association for Computational Linguistics.

\bibitem[{Lee et~al.(2022{\natexlab{b}})Lee, Gong, Duquenne, Schwenk, Chen, Wang, Popuri, Adi, Pino, Gu, and Hsu}]{lee-etal-2022-textless}
Ann Lee, Hongyu Gong, Paul-Ambroise Duquenne, Holger Schwenk, Peng-Jen Chen, Changhan Wang, Sravya Popuri, Yossi Adi, Juan Pino, Jiatao Gu, and Wei-Ning Hsu. 2022{\natexlab{b}}.
\newblock \href {https://doi.org/10.18653/v1/2022.naacl-main.63} {Textless speech-to-speech translation on real data}.
\newblock In \emph{Proceedings of the 2022 Conference of the North American Chapter of the Association for Computational Linguistics: Human Language Technologies}, pages 860--872, Seattle, United States. Association for Computational Linguistics.

\bibitem[{Morise et~al.(2016)Morise, Yokomori, and Ozawa}]{morise2016world}
Masanori Morise, Fumiya Yokomori, and Kenji Ozawa. 2016.
\newblock World: A vocoder-based high-quality speech synthesis system for real-time applications.
\newblock \emph{IEICE Trans. Inf. Syst.}, 99-D:1877--1884.

\bibitem[{Nakamura et~al.(2006)Nakamura, Markov, Nakaiwa, Kikui, Kawai, Jitsuhiro, Zhang, Yamamoto, Sumita, and Yamamoto}]{nakamura2006the}
S.~Nakamura, K.~Markov, H.~Nakaiwa, G.~Kikui, H.~Kawai, T.~Jitsuhiro, J.-S. Zhang, H.~Yamamoto, E.~Sumita, and S.~Yamamoto. 2006.
\newblock The atr multilingual speech-to-speech translation system.
\newblock \emph{IEEE Transactions on Audio, Speech, and Language Processing}, 14(2):365--376.

\bibitem[{Oreshkin et~al.(2018)Oreshkin, Rodriguez, and Lacoste}]{boris2018tadam}
Boris~N. Oreshkin, Pau Rodriguez, and Alexandre Lacoste. 2018.
\newblock Tadam: Task dependent adaptive metric for improved few-shot learning.
\newblock In \emph{Proceedings of the 32nd International Conference on Neural Information Processing Systems}, NIPS'18, page 719–729, Red Hook, NY, USA. Curran Associates Inc.

\bibitem[{Ott et~al.(2019)Ott, Edunov, Baevski, Fan, Gross, Ng, Grangier, and Auli}]{ott2019fairseq}
Myle Ott, Sergey Edunov, Alexei Baevski, Angela Fan, Sam Gross, Nathan Ng, David Grangier, and Michael Auli. 2019.
\newblock fairseq: A fast, extensible toolkit for sequence modeling.
\newblock In \emph{Proceedings of the 2019 Conference of the North {A}merican Chapter of the Association for Computational Linguistics (Demonstrations)}, pages 48--53.

\bibitem[{Pankov et~al.(2023)Pankov, Pronina, Kuzmin, Borisov, Usoltsev, Zeng, Golubkov, Ermolenko, Shirshova, and Matveeva}]{pankov2023dinovits}
Vikentii Pankov, Valeria Pronina, Alexander Kuzmin, Maksim Borisov, Nikita Usoltsev, Xingshan Zeng, Alexander Golubkov, Nikolai Ermolenko, Aleksandra Shirshova, and Yulia Matveeva. 2023.
\newblock \href {http://arxiv.org/abs/2311.09770} {Dino-vits: Data-efficient noise-robust zero-shot voice cloning via multi-tasking with self-supervised speaker verification loss}.

\bibitem[{Papineni et~al.(2002)Papineni, Roukos, Ward, and Zhu}]{papineni2002bleu}
Kishore Papineni, Salim Roukos, Todd Ward, and Wei-Jing Zhu. 2002.
\newblock \href {https://doi.org/10.3115/1073083.1073135} {{B}leu: a method for automatic evaluation of machine translation}.
\newblock In \emph{Proceedings of the 40th Annual Meeting of the Association for Computational Linguistics}, pages 311--318, Philadelphia, Pennsylvania, USA. Association for Computational Linguistics.

\bibitem[{Park et~al.(2019)Park, Chan, Zhang, Chiu, Zoph, Cubuk, and Le}]{park2019specaugment}
Daniel~S. Park, William Chan, Yu~Zhang, Chung-Cheng Chiu, Barret Zoph, Ekin~D. Cubuk, and Quoc~V. Le. 2019.
\newblock {SpecAugment: A Simple Data Augmentation Method for Automatic Speech Recognition}.
\newblock In \emph{Proc. Interspeech 2019}, pages 2613--2617.

\bibitem[{Perez et~al.(2018)Perez, Strub, de~Vries, Dumoulin, and Courville}]{perez2018film}
Ethan Perez, Florian Strub, Harm de~Vries, Vincent Dumoulin, and Aaron~C. Courville. 2018.
\newblock Film: Visual reasoning with a general conditioning layer.
\newblock In \emph{AAAI}.

\bibitem[{Popuri et~al.(2022)Popuri, Chen, Wang, Pino, Adi, Gu, Hsu, and Lee}]{popuri2022enhanced}
Sravya Popuri, Peng-Jen Chen, Changhan Wang, Juan Pino, Yossi Adi, Jiatao Gu, Wei-Ning Hsu, and Ann Lee. 2022.
\newblock Enhanced direct speech-to-speech translation using self-supervised pre-training and data augmentation.
\newblock \emph{arXiv preprint arXiv:2204.02967}.

\bibitem[{Radford et~al.(2022)Radford, Kim, Xu, Brockman, McLeavey, and Sutskever}]{whisper}
Alec Radford, Jong~Wook Kim, Tao Xu, Greg Brockman, Christine McLeavey, and Ilya Sutskever. 2022.
\newblock Robust speech recognition via large-scale weak supervision.
\newblock \emph{arXiv preprint arXiv:2212.04356}.

\bibitem[{Ren et~al.(2021)Ren, Hu, Qin, Zhao, Zhao, and Liu}]{ren2021fastspeech2}
Yi~Ren, Chenxu Hu, Tao Qin, Sheng Zhao, Zhou Zhao, and Tie-Yan Liu. 2021.
\newblock {FastSpeech} 2: Fast and high-quality end-to-end text-to-speech.
\newblock In \emph{Proc. ICLR}.

\bibitem[{Ren et~al.(2019)Ren, Ruan, Tan, Qin, Zhao, Zhao, and Liu}]{ren2019fastspeech}
Yi~Ren, Yangjun Ruan, Xu~Tan, Tao Qin, Sheng Zhao, Zhou Zhao, and Tie{-}Yan Liu. 2019.
\newblock Fast{S}peech: Fast, robust and controllable text to speech.
\newblock In \emph{Proc. NeurIPS}, pages 3165--3174.

\bibitem[{Roman et~al.(2024)Roman, Fernandez, Défossez, Furon, Tran, and Elsahar}]{roman2024proactive}
Robin~San Roman, Pierre Fernandez, Alexandre Défossez, Teddy Furon, Tuan Tran, and Hady Elsahar. 2024.
\newblock \href {http://arxiv.org/abs/2401.17264} {Proactive detection of voice cloning with localized watermarking}.

\bibitem[{Salimans and Kingma(2016)}]{tim2016weight}
Tim Salimans and Durk~P Kingma. 2016.
\newblock Weight normalization: A simple reparameterization to accelerate training of deep neural networks.
\newblock In \emph{Proc. NIPS}, pages 901--909.

\bibitem[{{Seamless Communication} et~al.(2023){Seamless Communication}, Barrault, Chung, Meglioli, Dale, Dong, Duquenne, Elsahar, Gong, Heffernan, Hoffman, Klaiber, Li, Licht, Maillard, Rakotoarison, Sadagopan, Wenzek, Ye, Akula, Chen, Hachem, Ellis, Gonzalez, Haaheim, Hansanti, Howes, Huang, Hwang, Inaguma, Jain, Kalbassi, Kallet, Kulikov, Lam, Li, Ma, Mavlyutov, Peloquin, Ramadan, Ramakrishnan, Sun, Tran, Tran, Tufanov, Vogeti, Wood, Yang, Yu, Andrews, Balioglu, Costa-jussà, Celebi, Elbayad, Gao, Guzmán, Kao, Lee, Mourachko, Pino, Popuri, Ropers, Saleem, Schwenk, Tomasello, Wang, Wang, and Wang}]{SeamlessM4TArXiv}
{Seamless Communication}, Loïc Barrault, Yu-An Chung, Mariano~Cora Meglioli, David Dale, Ning Dong, Paul-Ambroise Duquenne, Hady Elsahar, Hongyu Gong, Kevin Heffernan, John Hoffman, Christopher Klaiber, Pengwei Li, Daniel Licht, Jean Maillard, Alice Rakotoarison, Kaushik~Ram Sadagopan, Guillaume Wenzek, Ethan Ye, Bapi Akula, Peng-Jen Chen, Naji~El Hachem, Brian Ellis, Gabriel~Mejia Gonzalez, Justin Haaheim, Prangthip Hansanti, Russ Howes, Bernie Huang, Min-Jae Hwang, Hirofumi Inaguma, Somya Jain, Elahe Kalbassi, Amanda Kallet, Ilia Kulikov, Janice Lam, Daniel Li, Xutai Ma, Ruslan Mavlyutov, Benjamin Peloquin, Mohamed Ramadan, Abinesh Ramakrishnan, Anna Sun, Kevin Tran, Tuan Tran, Igor Tufanov, Vish Vogeti, Carleigh Wood, Yilin Yang, Bokai Yu, Pierre Andrews, Can Balioglu, Marta~R. Costa-jussà, Onur Celebi, Maha Elbayad, Cynthia Gao, Francisco Guzmán, Justine Kao, Ann Lee, Alexandre Mourachko, Juan Pino, Sravya Popuri, Christophe Ropers, Safiyyah Saleem, Holger Schwenk, Paden Tomasello, Changhan Wang, Jeff
  Wang, and Skyler Wang. 2023.
\newblock \href {https://arxiv.org/abs/2308.11596} {Seamlessm4t-massively multilingual \& multimodal machine translation}.

\bibitem[{{Seamless Communication}(2023)}]{barrault2023seamless}
Yu-An Chung Mariano Coria Meglioli David Dale Ning Dong Mark Duppenthaler Paul-Ambroise Duquenne Brian Ellis Hady Elsahar Justin Haaheim John Hoffman Min-Jae Hwang Hirofumi Inaguma Christopher Klaiber Ilia Kulikov Pengwei Li Daniel Licht Jean Maillard Ruslan Mavlyutov Alice Rakotoarison Kaushik Ram Sadagopan Abinesh Ramakrishnan Tuan Tran Guillaume Wenzek Yilin Yang Ethan Ye Ivan Evtimov Pierre Fernandez Cynthia Gao Prangthip Hansanti Elahe Kalbassi Amanda Kallet Artyom Kozhevnikov Gabriel Mejia Robin San Roman Christophe Touret Corinne Wong Carleigh Wood Bokai Yu Pierre Andrews Can Balioglu Peng-Jen Chen Marta R. Costa-juss{\`a} Maha Elbayad Hongyu Gong Francisco Guzm{\'a}n Kevin Heffernan Somya Jain Justine Kao Ann Lee Xutai Ma Alex Mourachko Benjamin Peloquin Juan Pino Sravya Popuri Christophe Ropers Safiyyah Saleem Holger Schwenk Anna Sun Paden Tomasello Changhan Wang Jeff Wang Skyler Wang Mary~Williamson {Seamless Communication}, Lo{\"i}c~Barrault. 2023.
\newblock Seamless: Multilingual expressive and streaming speech translation.

\bibitem[{Silero(2021)}]{SileroVAD}
Silero. 2021.
\newblock Silero vad: pre-trained enterprise-grade voice activity detector (vad), number detector and language classifier.
\newblock \url{https://github.com/snakers4/silero-vad}.

\bibitem[{Skerry-Ryan et~al.(2018)Skerry-Ryan, Battenberg, Xiao, Wang, Stanton, Shor, Weiss, Clark, and Saurous}]{skerry2018towards}
RJ~Skerry-Ryan, Eric Battenberg, Ying Xiao, Yuxuan Wang, Daisy Stanton, Joel Shor, Ron Weiss, Rob Clark, and Rif~A. Saurous. 2018.
\newblock \href {https://proceedings.mlr.press/v80/skerry-ryan18a.html} {Towards end-to-end prosody transfer for expressive speech synthesis with tacotron}.
\newblock In \emph{Proceedings of the 35th International Conference on Machine Learning}, volume~80 of \emph{Proceedings of Machine Learning Research}, pages 4693--4702. PMLR.

\bibitem[{Song et~al.(2023)Song, Ren, Lei, Wang, Wei, Xie, Yin, and Ma}]{song2023styles2st}
Kun Song, Yi~Ren, Yi~Lei, Chunfeng Wang, Kun Wei, Lei Xie, Xiang Yin, and Zejun Ma. 2023.
\newblock Styles2st: Zero-shot style transfer for direct speech-to-speech translation.
\newblock \emph{arXiv preprint arXiv:2305.17732}.

\bibitem[{Swiatkowski et~al.(2023)Swiatkowski, Wang, Babianski, Tobing, Vipperla, and Pollet}]{swiatkowski2023cross}
Jakub Swiatkowski, Duo Wang, Mikolaj Babianski, Patrick~Lumban Tobing, Ravichander Vipperla, and Vincent Pollet. 2023.
\newblock Cross-lingual prosody transfer for expressive machine dubbing.
\newblock \emph{arXiv preprint arXiv:2306.11658}.

\bibitem[{Valk and Alum{\"a}e(2021)}]{valk2021slt}
J{\"o}rgen Valk and Tanel Alum{\"a}e. 2021.
\newblock {VoxLingua107}: a dataset for spoken language recognition.
\newblock In \emph{Proc. IEEE SLT Workshop}.

\bibitem[{van~der Maaten and Hinton(2008)}]{van2008visualizing}
Laurens van~der Maaten and Geoffrey Hinton. 2008.
\newblock Visualizing data using {t-SNE}.
\newblock \emph{Journal of Machine Learning Research}, 9:2579--2605.

\bibitem[{Wahlster(2013)}]{wahlster2013verbmobil}
Wolfgang Wahlster. 2013.
\newblock \emph{Verbmobil: foundations of speech-to-speech translation}.
\newblock Springer Science \& Business Media.

\end{thebibliography}
\bibliographystyle{acl_natbib}

\appendix
\section{Benchmarking data}
\label{sec:appendix:benchmark_data}
\begin{table}[htbp!]
\footnotesize
\setlength\tabcolsep{3.pt}
\renewcommand{\arraystretch}{0.5}
\centering
\begin{tabular}{c|c|cc}
\toprule
Data & Subset & \# utterances & Duration (hrs) \\\midrule
\multirow{2}{*}{mExpresso (En$\rightarrow$Es)} & Dev & 4,758 & 4.17 \\
 & Test & 5,703 & 5.56 \\\midrule
\multirow{2}{*}{mDRAL (Es$\rightarrow$En)} & Dev & 587 & 0.46 \\
 & Test & 430 & 0.32\\ \midrule
DNS-5 & -- & 63,810 & 177.13 \\  
\bottomrule
\end{tabular}
\caption{Statistics of benchmarking datasets.}
\label{table:benchmark_data}
\end{table}
The statistics of benchmarking data are shown in \Cref{table:benchmark_data}.
The mExpresso En$\rightarrow$Es dataset is distributed under the CC-BY-NC 4.0 license.
The DNS-5 dataset is distributed under CC-BY-NC 0.0 and 4.0 licenses. 




\section{Pre-processing}
\label{sec:appendix:preprocessing}
All preprocessings were performed after adjusting sampling rate of input speech to 16-kHz.

\paragraph{Discrete units.}
Following \citet{SeamlessM4TArXiv}, we extracted continuous speech representations from 35$^{th}$ layer of XLS-R-1B model~\citep{babu2021xls} at 20-ms frame interval.
Then, we applied 10K K-means clustering algorithm on these representations to obtain discretized units.
Finally, we deduplicated it to have unique value at each unit sequence their duration.

\paragraph{Mel-spectrogram.}
We extracted 80-dimensional Mel-spectrograms with frame size and hop size of 400 (25-ms) and 160 (10-ms).
We applied zero-mean and unit-variance normalization to input and output Mel-filterbank features to stabilize model training.

\paragraph{Pitch.}
To extract F0 and VUV flag, we first extracted F0 in every 5-ms by using DIO algorithm~\citep{morise2016world}.
Then, we obtained VUV flag specifying non-zero values of F0, while obtaining continuous F0 contour by linearly interpolating zero values.
Finally, we converted linear F0 and energy values into log-scale.
Using the duration of unit, F0 and VUV flag features were averaged to have a reduced unit-scale.

\paragraph{Energy.}
To extract energy, we extracted energy contour every 5 ms using a 35-ms Hanning window.
Using the duration of unit, energy feature was averaged to have a reduced unit-scale.

\section{Architecture}
\label{sec:appendix:arch}
\begin{table*}[t]
\small
\centering

\begin{tabular}{l|l|c}
\toprule
                                                                                                     & \multicolumn{2}{c}{Hyperparemeter}                           \\ \midrule
\multirow{14}{*}{\begin{tabular}[c]{@{}l@{}}Teacher and student\\Expressivity encoders\end{tabular}} & Initial TDNN block hidden dimension              & 512                  \\
                                                                                                     & Initial TDNN block kernel size              & 5                    \\
                                                                                                     & SE-Res2Net block hidden dimension                & {[}512, 512, 512{]}  \\
                                                                                                     & SE-Res2Net block kernel size                & {[}3, 3, 3{]}        \\
                                                                                                     & SE-Res2Net block dilation              & {[}2, 3, 4{]}        \\
                                                                                                     & Res2Net scale                          & 8                    \\
                                                                                                     & Attentive statistic pooling hidden dimension     & 128                  \\
                                                                                                     & Last TDNN block hidden dimension                 & 1,536                 \\
                                                                                                     & Last TDNN block kernel size                 & 1                    \\
                                                                                                     & Expressivity embedding dimension       & 512                  \\
                                                                                                     & DINO projection layers                 & 3                    \\
                                                                                                     & DINO projection hidden dimension                 & 2,048                 \\
                                                                                                     & DINO projection bottleneck dimension      & 256                  \\
                                                                                                     & DINO projection output dimension       & 65,536               \\ \midrule
\multirow{19}{*}{Accoustic model}                                                                    & Unit embedding dim.                    & 256                  \\
                                                                                                     & Encoder layers                         & 4                    \\
                                                                                                     & Encoder hidden dimension                         & 256                  \\
                                                                                                     & Encoder Conv1D kernel size                   & 9                    \\
                                                                                                     & Encoder Conv1D channel                  & 1,024                 \\
                                                                                                     & Encoder attention heads                & 2                    \\
                                                                                                     & Local prosody predictor Conv1D kernel size  & 5                    \\
                                                                                                     & Local prosody predictor Conv1D channel & 512                  \\
                                                                                                     & Local prosody predictor dropout        & 0.5                  \\
                                                                                                     & Decoder layers                         & 4                    \\
                                                                                                     & Decoder hidden dimension                         & 256                  \\
                                                                                                     & Decoder Conv1D kernel size                  & 9                    \\
                                                                                                     & Decoder Conv1D channels                & 1,024                 \\
                                                                                                     & Decoder Conv1D attention heads         & 2                    \\
                                                                                                     & Encoder-decoder dropout                & 0.2                  \\
                                                                                                     & PostNet layers                         & 5                    \\
                                                                                                     & PostNet Conv1D channel                 & 512                  \\
                                                                                                     & PostNet Conv1D kernel size                  & 5                    \\
                                                                                                     & PostNet dropout                        & 0.5                  \\ \midrule
\multicolumn{2}{c|}{Total number of parameters for training}                                                                                                & \multicolumn{1}{c}{361M} \\
\multicolumn{2}{c|}{Total number of parameters for inference}                                                                                                & \multicolumn{1}{c}{67M}
\\ \bottomrule
\end{tabular}

\caption{Hyperparameters of \dinopretssel.}
\label{table:hyperparameters}
\end{table*}
The details of hyperparameter used for \dinopretssel architecture are described in \Cref{table:hyperparameters}.
The hyperparameters were selected based on the ones that performed the best in our experiments.
Note that the DINO training strategy requires an additional expressivity encoder with projection layers, which increases the required number of network parameters. 
However, these additional components are not necessary for inference, allowing for a reduction in model size.

\section{Training}
\label{sec:appendix:training}
For the other \pretssel models, we trained the model by $800$k iterations.
The loss coefficients for local prosody prediction $\lambda_{local}$, FiLM regularization $\lambda_{film}$, and DINO loss $\lambda_{dino}$ were set to $1.0$, $10^{-4}$, and $0.5$, respectively.
We used the Adam optimizer with $\beta_1 = 0.9$ and $\beta_2 = 0.98$ with fixed learning rate of $10^{-4}$.
We used gradient accumulation with frequency of four~\citep{hermans2017accumulated}.
Total 16 V100 GPUs were used to train a models.
The \dinopretssel and \pretssel spent total 12.7 and 7.9 days for their pretraining, respectively.
We implement our models based on the Fairseq toolkit \citep{ott2019fairseq}. 

In addition to the noise augmentation, we applied SpecAugment~\citep{park2019specaugment} with frequency mask with a maximum width of 8 and time mask with a maximum width of 10 during training.
Note that SpecAugment was only applied at student encoder.
To alleviate language imbalance in training data, we applied temperature-based resampling~\citep{ari2019thecurious} with the temperature set to 5.

For the DINO-related training setting,we adopted the commonly used DINO hyperparameters, which has proven to be robust in the works of image representation~\citep{caron2021emerging} speaker verification~\citep{chen2023comprehensive}, and text-to-speech~\citep{pankov2023dinovits}.
Specifically, we set the student temperature $\tau_s$ to 0.1, whereas we linearly increased the teacher temperature $\tau_t$ from 0.04 to 0.07 during first 20,000 iterations.
The momentum factor $m$ for EMA rule of mean statistic was set to 0.9.
EMA coefficients $\lambda_{ema}$ for teacher network update was initially set to 0.996, and gradually increased to 1.0 using cosine scheduler~\citep{grill2020bootstrap}. 

\section{Objective evaluation metrics}
\label{sec:appendix:obj_eval_metrics}
All metrics were measured after adjusting speech's sampling rate to 16-kHz.

\paragraph{ASR-BLEU.}
We obtained normalized transcription of translated speech using Whisper-Large ASR model\footnote{\url{https://github.com/openai/whisper}}~\citep{whisper}.
Then, we measured BLEU scores between reference and transcriptions~\citep{papineni2002bleu}.

\paragraph{S2T-BLEU.}
We obtained translated text from S2TT components of Prosody UnitY2.
Then, we measured BLEU scores between reference and transcriptions~\citep{papineni2002bleu}.

\paragraph{AutoPCP.}
We computed utterance-level prosody preservation score by using AutoPCP-multilingual-v2 model on source language and translated target language speeches~\citep{barrault2023seamless}.

\paragraph{SNR.}
Following the work of \citet{barrault2023seamless}, we computed SNR from the energy ratio between denoised speech and residual speech.
Specifically, we applied DEMUCs denoiser\footnote{\url{https://github.com/facebookresearch/denoiser}}~\citep{defossez2020real} to the original speech $\mathbf{s}$ for obtaining denoised speech $\hat{\mathbf{s}}$. 
Then, we computed SNR as follows:
\begin{equation}
SNR = 10\text{log}_{10} \left(\frac{||\hat{\mathbf{s}}||_2^2}{||\mathbf{s} - \hat{\mathbf{s}}||_2^2}\right),
\end{equation}
where $||\cdot||_2$ denotes L2 norm.

\section{Detailed objective evaluation scores}
\label{sec:appendix:detailed_obj_eval_scores}
\paragraph{Evaluation of S2ST system with predicted units.}

\begin{table*}[t]
\footnotesize
\setlength\tabcolsep{3pt}
\centering
\begin{tabular}{c|c|l|cccc}
\toprule
Subset & SNR (dB) & \multicolumn{1}{c|}{System} & \multicolumn{1}{c}{S2T-BLEU$\uparrow$} & \multicolumn{1}{c}{ASR-BLEU$\uparrow$} & \multicolumn{1}{c}{AutoPCP$\uparrow$} & \multicolumn{1}{c}{Estimated SNR (dB)$\uparrow$} \\ \midrule

\multirow{24}{*}{\begin{tabular}[c]{@{}c@{}}Dev\end{tabular}} & \multirow{4}{*}{Clean} & S2TT + TTS & 46.18 & 37.59 & 3.02 & 28.84 \\
 &  & PRETSSEL & 46.18 & 41.45 & 3.46 & 28.41 \\
 &  & Denoiser + PRETSSEL & 46.18 & 41.44 & 3.34 & 33.45 \\
 &  & DINO-PRETSSEL (proposed) & 46.18 & 41.56 & 3.36 & 29.46 \\\cmidrule{2-7} 
 & \multirow{4}{*}{0} & S2TT + TTS & 36.65 & 23.46 & 2.30 & 11.71 \\
 &  & PRETSSEL & 36.65 & 30.28 & 2.45 & 5.20 \\
 &  & Denoiser + PRETSSEL & 36.65 & 32.40 & 2.84 & 12.87 \\
 &  & DINO-PRETSSEL (proposed) & 36.65 & 33.61 & 2.94 & 21.08 \\\cmidrule{2-7} 
 & \multirow{4}{*}{5} & S2TT + TTS & 40.36 & 27.74 & 2.46 & 14.82 \\
 &  & PRETSSEL & 40.36 & 35.14 & 2.73 & 8.91 \\
 &  & Denoiser + PRETSSEL & 40.36 & 36.14 & 3.00 & 21.94 \\
 &  & DINO-PRETSSEL (proposed) & 40.36 & 37.10 & 3.05 & 24.45 \\\cmidrule{2-7} 
 & \multirow{4}{*}{10} & S2TT + TTS & 42.64 & 31.55 & 2.60 & 17.38 \\
 &  & PRETSSEL & 42.64 & 37.56 & 2.95 & 13.37 \\
 &  & Denoiser + PRETSSEL & 42.64 & 38.55 & 3.11 & 28.20 \\
 &  & DINO-PRETSSEL (proposed) & 42.64 & 39.04 & 3.13 & 26.71 \\\cmidrule{2-7} 
 & \multirow{4}{*}{15} & S2TT + TTS & 43.87 & 34.02 & 2.71 & 19.90 \\
 &  & PRETSSEL & 43.87 & 39.07 & 3.10 & 17.78 \\
 &  & Denoiser + PRETSSEL & 43.87 & 39.51 & 3.19 & 31.50 \\
 &  & DINO-PRETSSEL (proposed) & 43.87 & 39.92 & 3.21 & 28.02 \\\cmidrule{2-7} 
 & \multirow{4}{*}{20} & S2TT + TTS & 44.95 & 35.61 & 2.80 & 22.33 \\
 &  & PRETSSEL & 44.95 & 40.19 & 3.22 & 21.62 \\
 &  & Denoiser + PRETSSEL & 44.95 & 40.51 & 3.26 & 32.75 \\
 &  & DINO-PRETSSEL (proposed) & 44.95 & 40.91 & 3.26 & 28.50 \\\midrule

\multirow{24}{*}{\begin{tabular}[c]{@{}c@{}}Test\end{tabular}} & \multirow{4}{*}{Clean} & S2TT + TTS & 47.00 & 38.68 & 2.89 & 28.15 \\
 &  & PRETSSEL & 47.00 & 42.30 & 3.30 & 28.40 \\
 &  & Denoiser + PRETSSEL & 47.00 & 42.13 & 3.16 & 33.42 \\
 &  & DINO-PRETSSEL (proposed) & 47.00 & 42.36 & 3.23 & 28.93 \\\cmidrule{2-7} 
 & \multirow{4}{*}{0} & S2TT + TTS & 38.70 & 24.22 & 2.24 & 12.24 \\
 &  & PRETSSEL & 38.70 & 32.35 & 2.27 & 4.78 \\
 &  & Denoiser + PRETSSEL & 38.70 & 34.16 & 2.69 & 14.31 \\
 &  & DINO-PRETSSEL (proposed) & 38.70 & 34.99 & 2.79 & 21.25 \\\cmidrule{2-7} 
 & \multirow{4}{*}{5} & S2TT + TTS & 43.03 & 30.35 & 2.39 & 15.32 \\
 &  & PRETSSEL & 43.03 & 37.28 & 2.56 & 8.95 \\
 &  & Denoiser + PRETSSEL & 43.03 & 38.27 & 2.85 & 23.29 \\
 &  & DINO-PRETSSEL (proposed) & 43.03 & 39.08 & 2.92 & 25.23 \\\cmidrule{2-7} 
 & \multirow{4}{*}{10} & S2TT + TTS & 44.94 & 34.04 & 2.50 & 18.10 \\
 &  & PRETSSEL & 44.94 & 39.41 & 2.78 & 13.29 \\
 &  & Denoiser + PRETSSEL & 44.94 & 40.12 & 2.96 & 28.91 \\
 &  & DINO-PRETSSEL (proposed) & 44.94 & 40.66 & 3.02 & 27.34 \\\cmidrule{2-7} 
 & \multirow{4}{*}{15} & S2TT + TTS & 45.95 & 35.74 & 2.60 & 20.43 \\
 &  & PRETSSEL & 45.95 & 40.46 & 2.94 & 17.52 \\
 &  & Denoiser + PRETSSEL & 45.95 & 40.94 & 3.05 & 31.46 \\
 &  & DINO-PRETSSEL (proposed) & 45.95 & 41.34 & 3.09 & 28.34 \\\cmidrule{2-7} 
 & \multirow{4}{*}{20} & S2TT + TTS & 46.39 & 37.47 & 2.69 & 22.52 \\
 &  & PRETSSEL & 46.39 & 41.11 & 3.06 & 21.33 \\
 &  & Denoiser + PRETSSEL & 46.39 & 41.35 & 3.11 & 32.71 \\
 &  & DINO-PRETSSEL (proposed) & 46.39 & 41.83 & 3.15 & 28.76 \\\bottomrule
\end{tabular}
\caption{Full objective evaluation results of mExpresso (En$\rightarrow$Es) in S2ST system.}
\label{table:obj_eval_full_results_mexpresso}
\end{table*}

\begin{table*}[t]
\footnotesize
\setlength\tabcolsep{3pt}
\centering
\begin{tabular}{c|c|l|cccc}
\toprule
Subset & SNR (dB) & \multicolumn{1}{c|}{System} & \multicolumn{1}{c}{S2T-BLEU$\uparrow$} & \multicolumn{1}{c}{ASR-BLEU$\uparrow$} & \multicolumn{1}{c}{AutoPCP$\uparrow$} & \multicolumn{1}{c}{Estimated SNR (dB)$\uparrow$} \\ \midrule
\multirow{24}{*}{\begin{tabular}[c]{@{}c@{}}Dev\end{tabular}} & \multirow{4}{*}{Clean} & S2TT + TTS & 53.66 & 38.97 & 2.73 & 17.84 \\
 &  & PRETSSEL & 53.66 & 51.68 & 3.18 & 21.63 \\
 &  & Denoiser + PRETSSEL & 53.66 & 51.31 & 3.16 & 29.39 \\
 &  & DINO-PRETSSEL (proposed) & 53.66 & 52.18 & 3.10 & 23.48 \\\cmidrule{2-7} 
 & \multirow{4}{*}{0} & S2TT + TTS & 42.89 & 22.94 & 2.04 & 10.93 \\
 &  & PRETSSEL & 42.89 & 40.66 & 2.23 & 3.72 \\
 &  & Denoiser + PRETSSEL & 42.89 & 42.25 & 2.62 & 8.09 \\
 &  & DINO-PRETSSEL (proposed) & 42.89 & 42.54 & 2.79 & 19.17 \\\cmidrule{2-7} 
 & \multirow{4}{*}{5} & S2TT + TTS & 48.64 & 30.27 & 2.18 & 12.74 \\
 &  & PRETSSEL & 48.64 & 46.64 & 2.49 & 7.74 \\
 &  & Denoiser + PRETSSEL & 48.64 & 47.43 & 2.76 & 16.62 \\
 &  & DINO-PRETSSEL (proposed) & 48.64 & 48.05 & 2.88 & 21.27 \\\cmidrule{2-7} 
 & \multirow{4}{*}{10} & S2TT + TTS & 49.66 & 37.92 & 2.42 & 15.66 \\
 &  & PRETSSEL & 49.66 & 47.04 & 2.68 & 11.37 \\
 &  & Denoiser + PRETSSEL & 49.66 & 49.26 & 2.90 & 23.12 \\
 &  & DINO-PRETSSEL (proposed) & 49.66 & 49.25 & 2.96 & 21.51 \\\cmidrule{2-7} 
 & \multirow{4}{*}{15} & S2TT + TTS & 51.42 & 39.76 & 2.49 & 16.19 \\
 &  & PRETSSEL & 51.42 & 50.37 & 2.84 & 14.70 \\
 &  & Denoiser + PRETSSEL & 51.42 & 50.70 & 2.98 & 26.57 \\
 &  & DINO-PRETSSEL (proposed) & 51.42 & 50.90 & 3.00 & 22.10 \\\cmidrule{2-7} 
 & \multirow{4}{*}{20} & S2TT + TTS & 52.24 & 37.78 & 2.45 & 16.22 \\
 &  & PRETSSEL & 52.24 & 51.13 & 2.95 & 17.18 \\
 &  & Denoiser + PRETSSEL & 52.24 & 51.34 & 3.04 & 29.11 \\
 &  & DINO-PRETSSEL (proposed) & 52.24 & 51.33 & 3.04 & 22.02 \\\midrule
 
\multirow{24}{*}{\begin{tabular}[c]{@{}c@{}}Test\end{tabular}} & \multirow{4}{*}{Clean} & S2TT + TTS & 53.82 & 46.57 & 2.75 & 22.07 \\
 &  & PRETSSEL & 53.82 & 50.92 & 3.18 & 30.34 \\
 &  & Denoiser + PRETSSEL & 53.82 & 51.14 & 3.17 & 35.25 \\
 &  & DINO-PRETSSEL (proposed) & 53.82 & 52.32 & 3.09 & 31.15 \\\cmidrule{2-7} 
 & \multirow{4}{*}{0} & S2TT + TTS & 38.17 & 18.11 & 1.94 & 11.91 \\
 &  & PRETSSEL & 38.17 & 35.89 & 2.23 & 4.45 \\
 &  & Denoiser + PRETSSEL & 38.17 & 34.30 & 2.59 & 7.79 \\
 &  & DINO-PRETSSEL (proposed) & 38.17 & 34.54 & 2.78 & 23.92 \\\cmidrule{2-7} 
 & \multirow{4}{*}{5} & S2TT + TTS & 44.94 & 30.73 & 2.41 & 16.61 \\
 &  & PRETSSEL & 44.94 & 41.73 & 2.48 & 9.12 \\
 &  & Denoiser + PRETSSEL & 44.94 & 40.85 & 2.77 & 18.57 \\
 &  & DINO-PRETSSEL (proposed) & 44.94 & 43.81 & 2.91 & 26.17 \\\cmidrule{2-7} 
 & \multirow{4}{*}{10} & S2TT + TTS & 49.36 & 33.59 & 2.20 & 16.06 \\
 &  & PRETSSEL & 49.36 & 46.93 & 2.66 & 13.87 \\
 &  & Denoiser + PRETSSEL & 49.36 & 47.89 & 2.89 & 26.69 \\
 &  & DINO-PRETSSEL (proposed) & 49.36 & 48.53 & 2.96 & 27.61 \\\cmidrule{2-7} 
 & \multirow{4}{*}{15} & S2TT + TTS & 51.60 & 46.83 & 2.53 & 19.17 \\
 &  & PRETSSEL & 51.60 & 50.22 & 2.81 & 18.24 \\
 &  & Denoiser + PRETSSEL & 51.60 & 50.33 & 2.99 & 30.51 \\
 &  & DINO-PRETSSEL (proposed) & 51.60 & 50.83 & 3.01 & 28.52 \\\cmidrule{2-7} 
 & \multirow{4}{*}{20} & S2TT + TTS & 53.35 & 47.44 & 2.56 & 20.51 \\
 &  & PRETSSEL & 53.35 & 51.80 & 2.93 & 22.06 \\
 &  & Denoiser + PRETSSEL & 53.35 & 52.23 & 3.06 & 32.51 \\
 &  & DINO-PRETSSEL (proposed) & 53.35 & 52.65 & 3.04 & 28.81 \\\bottomrule
\end{tabular}
\caption{Full objective evaluation results of mDRAL (Es$\rightarrow$En) in S2ST system.}
\label{table:obj_eval_full_results_mdral}
\end{table*}

The detailed results of S2ST systems for mExpresso (En$\rightarrow$Es) and mDRAL (Es$\rightarrow$En) are shown in \Cref{table:obj_eval_full_results_mexpresso} and \Cref{table:obj_eval_full_results_mdral}, respectively.

\paragraph{Evaluation of U2S models with ground-truth units.}

\begin{table*}[t]
\footnotesize
\setlength\tabcolsep{3pt}
\centering
\centering
\begin{tabular}{c|c|c|ccc}
\toprule
Subset & SNR (dB) & System & ASR-BLEU$\uparrow$ & AutoPCP$\uparrow$ & Estimated SNR (dB)$\uparrow$ \\ \midrule

\multirow{18}{*}{Dev} & \multirow{3}{*}{Clean} & PRETSSEL & 79.48 & 3.45 & 27.79 \\
 &  & Denoiser + PRETSSEL & 79.48 & 3.35 & 33.00 \\
 &  & DINO-PRETSSEL (proposed) & 79.78 & 3.35 & 28.26 \\\cmidrule{2-6} 
 & \multirow{3}{*}{0} & PRETSSEL & 71.62 & 2.66 & 4.98 \\
 &  & Denoiser + PRETSSEL & 75.74 & 3.04 & 12.63 \\
 &  & DINO-PRETSSEL (proposed) & 79.28 & 3.14 & 19.91 \\\cmidrule{2-6} 
 & \multirow{3}{*}{5} & PRETSSEL & 74.93 & 2.88 & 8.81 \\
 &  & Denoiser + PRETSSEL & 77.44 & 3.15 & 21.06 \\
 &  & DINO-PRETSSEL (proposed) & 79.33 & 3.21 & 22.56 \\\cmidrule{2-6} 
 & \multirow{3}{*}{10} & PRETSSEL & 77.11 & 3.07 & 13.11 \\
 &  & Denoiser + PRETSSEL & 78.33 & 3.23 & 26.64 \\
 &  & DINO-PRETSSEL (proposed) & 79.35 & 3.25 & 24.39 \\\cmidrule{2-6} 
 & \multirow{3}{*}{15} & PRETSSEL & 78.15 & 3.19 & 17.26 \\
 &  & Denoiser + PRETSSEL & 79.11 & 3.27 & 29.65 \\
 &  & DINO-PRETSSEL (proposed) & 79.45 & 3.28 & 25.63 \\\cmidrule{2-6} 
 & \multirow{3}{*}{20} & PRETSSEL & 78.51 & 3.27 & 20.89 \\
 &  & Denoiser + PRETSSEL & 79.01 & 3.31 & 31.22 \\
 &  & DINO-PRETSSEL (proposed) & 79.64 & 3.30 & 26.36 \\ \midrule

\multirow{18}{*}{Test} & \multirow{3}{*}{Clean} & PRETSSEL & 79.89 & 3.33 & 26.69 \\
 &  & Denoiser + PRETSSEL & 79.93 & 3.20 & 32.19 \\
 &  & DINO-PRETSSEL (proposed) & 80.38 & 3.23 & 26.63 \\\cmidrule{2-6} 
 & \multirow{3}{*}{0} & PRETSSEL & 70.64 & 2.52 & 4.80 \\
 &  & Denoiser + PRETSSEL & 76.70 & 2.89 & 14.63 \\
 &  & DINO-PRETSSEL (proposed) & 80.09 & 3.03 & 20.10 \\\cmidrule{2-6} 
 & \multirow{3}{*}{5} & PRETSSEL & 75.38 & 2.75 & 8.91 \\
 &  & Denoiser + PRETSSEL & 78.16 & 3.00 & 22.48 \\
 &  & DINO-PRETSSEL (proposed) & 80.51 & 3.10 & 22.96 \\\cmidrule{2-6} 
 & \multirow{3}{*}{10} & PRETSSEL & 77.65 & 2.93 & 13.09 \\
 &  & Denoiser + PRETSSEL & 79.11 & 3.08 & 27.43 \\
 &  & DINO-PRETSSEL (proposed) & 80.59 & 3.14 & 24.61 \\\cmidrule{2-6} 
 & \multirow{3}{*}{15} & PRETSSEL & 78.58 & 3.05 & 17.18 \\
 &  & Denoiser + PRETSSEL & 79.45 & 3.13 & 29.98 \\
 &  & DINO-PRETSSEL (proposed) & 80.80 & 3.17 & 25.54 \\\cmidrule{2-6} 
 & \multirow{3}{*}{20} & PRETSSEL & 79.08 & 3.13 & 20.83 \\
 &  & Denoiser + PRETSSEL & 79.58 & 3.16 & 31.30 \\
 &  & DINO-PRETSSEL (proposed) & 80.75 & 3.19 & 25.98 \\\bottomrule
\end{tabular}
\caption{Full objective evaluation results of mExpresso (En$\rightarrow$Es) with ground-truth units.}
\label{table:obj_eval_full_results_gold_mexpresso}
\end{table*}

\begin{table*}[t]
\footnotesize
\setlength\tabcolsep{3pt}
\centering
\centering
\begin{tabular}{c|c|c|ccc}
\toprule
Subset & SNR (dB) & System & ASR-BLEU$\uparrow$ & AutoPCP$\uparrow$ & Estimated SNR (dB)$\uparrow$ \\ \midrule
\multirow{18}{*}{Dev} & \multirow{3}{*}{Clean} & PRETSSEL & 77.70 & 3.21 & 20.74 \\
 &  & Denoiser + PRETSSEL & 78.97 & 3.16 & 27.15 \\
 &  & DINO-PRETSSEL (proposed) & 79.50 & 3.09 & 21.48 \\\cmidrule{2-6} 
 & \multirow{3}{*}{0} & PRETSSEL & 63.95 & 2.35 & 3.48 \\
 &  & Denoiser + PRETSSEL & 71.84 & 2.70 & 6.53 \\
 &  & DINO-PRETSSEL (proposed) & 81.09 & 2.95 & 17.38 \\\cmidrule{2-6} 
 & \multirow{3}{*}{5} & PRETSSEL & 70.52 & 2.54 & 7.69 \\
 &  & Denoiser + PRETSSEL & 77.15 & 2.82 & 15.00 \\
 &  & DINO-PRETSSEL (proposed) & 78.95 & 3.01 & 18.48 \\\cmidrule{2-6} 
 & \multirow{3}{*}{10} & PRETSSEL & 73.12 & 2.71 & 11.29 \\
 &  & Denoiser + PRETSSEL & 78.69 & 2.92 & 21.47 \\
 &  & DINO-PRETSSEL (proposed) & 81.18 & 3.03 & 19.47 \\\cmidrule{2-6} 
 & \multirow{3}{*}{15} & PRETSSEL & 76.31 & 2.87 & 14.37 \\
 &  & Denoiser + PRETSSEL & 78.02 & 3.00 & 25.04 \\
 &  & DINO-PRETSSEL (proposed) & 79.66 & 3.05 & 20.02 \\\cmidrule{2-6} 
 & \multirow{3}{*}{20} & PRETSSEL & 75.89 & 2.97 & 16.51 \\
 &  & Denoiser + PRETSSEL & 78.37 & 3.04 & 26.98 \\
 &  & DINO-PRETSSEL (proposed) & 79.94 & 3.06 & 20.18 \\ \midrule

\multirow{18}{*}{Test} & \multirow{3}{*}{Clean} & PRETSSEL & 78.13 & 3.20 & 28.46 \\
 &  & Denoiser + PRETSSEL & 78.67 & 3.20 & 32.82 \\
 &  & DINO-PRETSSEL (proposed) & 78.86 & 3.09 & 28.78 \\\cmidrule{2-6} 
 & \multirow{3}{*}{0} & PRETSSEL & 60.03 & 2.30 & 4.02 \\
 &  & Denoiser + PRETSSEL & 70.37 & 2.67 & 6.56 \\
 &  & DINO-PRETSSEL (proposed) & 79.97 & 2.94 & 23.15 \\\cmidrule{2-6} 
 & \multirow{3}{*}{5} & PRETSSEL & 69.75 & 2.51 & 8.76 \\
 &  & Denoiser + PRETSSEL & 75.92 & 2.79 & 17.43 \\
 &  & DINO-PRETSSEL (proposed) & 80.53 & 2.99 & 25.03 \\\cmidrule{2-6} 
 & \multirow{3}{*}{10} & PRETSSEL & 72.14 & 2.68 & 13.32 \\
 &  & Denoiser + PRETSSEL & 78.23 & 2.91 & 24.70 \\
 &  & DINO-PRETSSEL (proposed) & 80.45 & 3.02 & 26.25 \\\cmidrule{2-6} 
 & \multirow{3}{*}{15} & PRETSSEL & 75.31 & 2.80 & 17.35 \\
 &  & Denoiser + PRETSSEL & 78.63 & 2.99 & 29.25 \\
 &  & DINO-PRETSSEL (proposed) & 79.56 & 3.03 & 26.69 \\\cmidrule{2-6} 
 & \multirow{3}{*}{20} & PRETSSEL & 76.09 & 2.91 & 20.96 \\
 &  & Denoiser + PRETSSEL & 78.91 & 3.05 & 31.31 \\
 &  & DINO-PRETSSEL (proposed) & 78.94 & 3.04 & 27.46 \\ \bottomrule
\end{tabular}
\caption{Full objective evaluation results mDRAL (Es$\rightarrow$En) with ground-truth units.}
\label{table:obj_eval_full_results_gold_mdral}
\end{table*}

The detailed results of U2S systems with ground-truth units for mExpresso (En$\rightarrow$Es) and mDRAL (Es$\rightarrow$En) are shown in \Cref{table:obj_eval_full_results_gold_mexpresso} and \Cref{table:obj_eval_full_results_gold_mdral}, respectively.

\section{Subjective evaluation metrics}
\label{sec:appendix:subj_eval_metrics}
For subjective evaluation, we randomly sampled total 100 and 90 utterances from the combined dev/test sets of mExpresso and mDRAL dataset, respectively.
In case of VoxLingua107 dataset, we randomly sampled total 100 utterances from its training set.
Each item was evaluated by three annotators.
Before conducting evaluation, they were informed on the purpose of the human evaluation studies.
A more detailed protocol explanation can be found in \citet{SeamlessM4TArXiv}.

\paragraph{MOS.}
We adopted the 5-point Likert scale MOS protocol~\citep{p808} to evaluate the speech quality.
The target language's native speakers were asked to rate speech's naturalness on the scores ranging from \textit{1. Extremely unnatural} to \textit{5. Extremely natural}.

\paragraph{S-MOS.}
We adopted an S-MOS protocol to measure the similarity of source- and target-voices.
Monolingual English listeners were asked to listen to both source and target audio and rate the similarity of the voices (disregarding the content and manner of the utterances) on a 5-point Likert scale ranging from \textit{1. Not at all similar} to \textit{5. Extremely similar}. 

\end{document}